%% file: acl_latex.tex
\documentclass[11pt]{article}

\usepackage[preprint]{acl}

\usepackage{times}
\usepackage{latexsym}

\usepackage[T1]{fontenc}
\usepackage[utf8]{inputenc}

\usepackage{microtype}

\usepackage{inconsolata}

\colorlet{soulblue}{blue!20}
\usepackage{graphicx}
\usepackage{longtable}
\usepackage{booktabs}
\usepackage{array}
\usepackage{xcolor, soul}
\usepackage{multirow}
\usepackage{diagbox}
\usepackage{amsmath}
\usepackage{enumitem}
\usepackage[most]{tcolorbox}

\tcbuselibrary{skins, breakable}

\usepackage{makecell}

\newcommand{\stat}[3]{%
    \makecell[c]{#1 \\ \textcolor{gray}{\scriptsize [#2--#3]}}%
}

\title{Scaling Therapy Supervision: Assessing Psychotherapy Fidelity Using Large Language Models}

\author{
  \textbf{Arvind Pillai\textsuperscript{1}},
  \textbf{Alexis Roach\textsuperscript{2}},
  \textbf{Jason Holden\textsuperscript{2}},
  \textbf{Varsha D Badal\textsuperscript{2}},
  \\
  \textbf{Subigya Nepal\textsuperscript{3}},
  \textbf{Yu Yvonne Wu\textsuperscript{1}},
  \textbf{Yuliang Chen\textsuperscript{1}},
  \textbf{Sudarshan Regmi \textsuperscript{1}},
  \\
  \textbf{Sarah Masud Preum \textsuperscript{1}},
  \textbf{Nicholas C Jacobson \textsuperscript{1}},
  \textbf{Nikhil Singh\textsuperscript{1}},
  \\
  \textbf{Eric Granholm \textsuperscript{1}},
  \textbf{Colin Depp\textsuperscript{1}},
  \textbf{Andrew Campbell\textsuperscript{1}},
  \\
  \textsuperscript{1}Dartmouth College,
  \textsuperscript{2}University of California, San Diego,
  \textsuperscript{3}University of Virginia,
  \\
  \small{
    \textbf{Correspondence:} \texttt{arvind.pillai.gr@dartmouth.edu}
  }
}

\begin{document}
\maketitle
\begin{abstract}
Psychotherapy fidelity assessment measures how faithfully a therapist adheres to a treatment protocol, a critical determinant of positive patient outcomes. However, fidelity assessment is bottlenecked by manual expert review, rendering the process labor-intensive and unscalable. To scale therapy supervision and support clinician self-learning, we present an integrated application of large language models (LLMs) that automates psychotherapy fidelity workflows and generates actionable supervisory feedback from authentic, long-form clinical transcripts. We evaluate three open-weight models across multiple task decomposition configurations and reasoning strategies, demonstrating that evidence-first reasoning improves fidelity estimation. Furthermore, we establish the framework's external construct validity on an independent randomized controlled trial (RCT), where it reliably distinguishes Cognitive Behavioral Therapy (CBT) from non-CBT sessions across core therapeutic skills. Through a user study that evaluates the generated feedback, mental health experts indicate that the pipeline synthesizes actionable clinical feedback. Ultimately, our work offers a practical, scalable solution toward bridging the community mental health supervision gap.

\end{abstract}

\section{Introduction}

\begin{figure}[htbp]
  \begin{center}
    \includegraphics[width=0.48\textwidth]{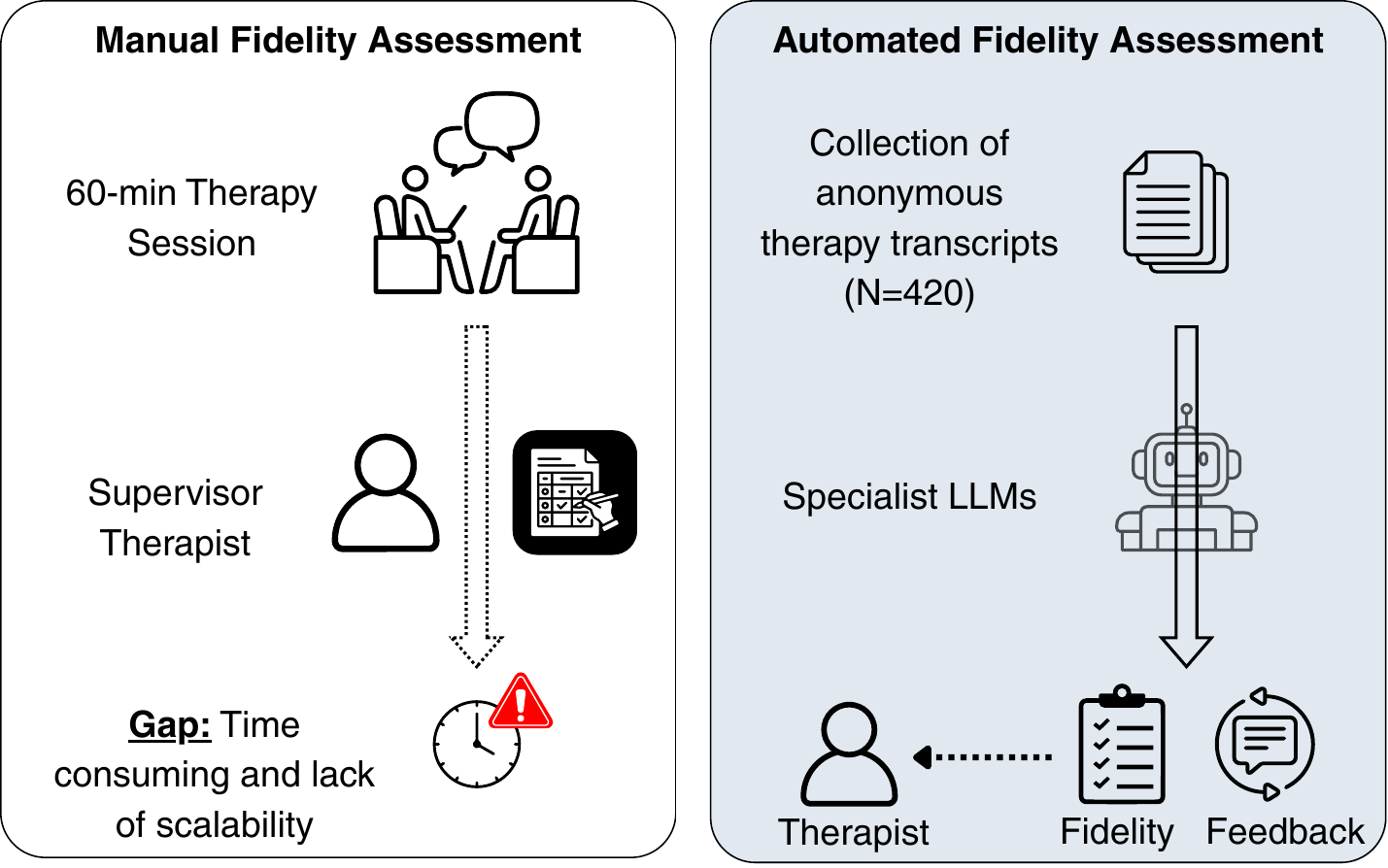}
  \end{center}
  \caption{\textbf{Scaling therapy supervision.} Left: Manual review of a 60-minute session is highly time-intensive, limiting scalability. Right: The proposed automated pipeline leverages LLM reasoning over de-identified transcripts to automatically generate fidelity ratings and actionable feedback.}
\label{fig:rq1}
\end{figure}

Serious mental illness (SMI) affects approximately 14.6 million people in the United States, with prevalence rising sharply among young adults, particularly for depression and psychosis~\citep{apa2019mentalhealth}. At the same time, mental health systems are under increasing strain: over half of clinical practices report no availability for new patients~\citep{apa2023psychologistlimits}. This capacity gap is compounded by the high labor cost of delivering and monitoring evidence-based therapy, since even gold-standard treatments depend on consistent, high-fidelity delivery that is currently difficult to sustain at scale.

Evidence-based psychotherapies for SMI, such as Cognitive Behavioral Therapy (CBT), rely on structured protocols, and adherence is one of the strongest predictors of positive patient outcomes~\citep{roberts2004evidence, spring2007evidence, cook2017evidence}. \textit{Fidelity} is how faithfully and competently a therapist delivers the treatment manual. For psychosis, it is measured using the Cognitive Therapy Scale for Psychosis (CTS-Psy)~\citep{haddock2001investigation}. However, fidelity assessment is a labor-intensive process requiring trained experts: reviewing a single session often takes longer than the session itself~\citep{waltman2017cognitive} (Figure \ref{fig:rq1}). As a result, systematic fidelity assessment is rarely performed at scale in community settings, depriving clinicians of the timely, performance-based feedback needed for training and reducing the burden on expert supervisors.

LLMs offer a path to closing this supervision gap. Prior automated fidelity work is dominated by transformer classifiers and feature-engineered models~\citep{creed2022enhancing, stanco2021lyssn, flemotomos2022automated, zainal2025capitalizing}. In contrast, LLMs can reason over an entire de-identified transcript in natural language, apply rubric criteria item-by-item, and return both a rating and supporting evidence, augmenting a human supervisor's judgment at a fraction of the time and cost. This has motivated a growing line of work that operationalizes clinical rubrics as LLM evaluation pipelines, typically adapting the Cognitive Therapy Rating Scale (CTRS) to score CBT adherence and benchmark model outputs against human clinicians~\citep{huang2026therapygym, chiu2024computational, bedi2026assessing, teferra2026alignment, lee2024cactus, rollwage2026cognitive}, with recent results suggesting LLMs can approach or exceed clinician-level competence at scale~\citep{rollwage2026cognitive}. Together, this evidence suggests that rubric-level clinical reasoning is within reach for current models.

% LLMs offer a path to closing this supervision gap. Prior automated fidelity work is dominated by transformer classifiers and feature-engineered models~\citep{creed2022enhancing, stanco2021lyssn, flemotomos2022automated, zainal2025capitalizing}. In contrast, LLMs can reason over an entire de-identified transcript in natural language, apply rubric criteria item-by-item, and return both a rating and supporting evidence. This augments the judgment of a human supervisor at a fraction of the time and cost. Such capabilities have motivated recent work that operationalizes clinical rubrics as LLM evaluation pipelines~\citep{huang2026therapygym, chiu2024computational}. Most of this work adapts the Cognitive Therapy Rating Scale (CTRS) to score adherence to CBT techniques and to assess how closely model outputs track human clinicians~\citep{bedi2026assessing, teferra2026alignment, lee2024cactus}. Some studies further use these scores as reward signals for training therapy models~\citep{huang2026therapygym}, and others report that LLMs can approach or exceed clinician-level competence at scale~\citep{rollwage2026cognitive}. Together, this evidence suggests that rubric-level clinical reasoning is within reach for current models.

However, this body of work overwhelmingly evaluates LLMs in the role of the therapist rather than as an assessment tool for real clinical sessions. It remains unclear whether the same models can serve as a practical, therapist-facing tool that assesses authentic, expert-rated human sessions, where transcripts are noisy and run far longer than a typical chat. In this work, we take an application-first stance: rather than proposing a new architecture, we ask whether existing open-weight LLMs can support real fidelity-assessment workflows.

% However, this body of work overwhelmingly evaluates LLMs in the role of the therapist, scoring synthetic, role-played, or model-generated dialogue~\citep{bedi2026assessing, chiu2024computational, huang2026therapygym, rollwage2026cognitive}. It remains unclear whether the same models can serve as a practical, therapist-facing tool that assesses authentic, expert-rated human sessions, where transcripts are noisy and run far longer than a typical chat. In this work, we take an application-first stance: rather than proposing a new architecture, we ask whether existing open-weight LLMs can support real fidelity-assessment workflows.

Toward this goal, our contributions are as follows. \textbf{(1) Automated fidelity assessment:} We systematically evaluate three open-weight LLMs on automated fidelity assessment across 420 authentic, long-form CBT transcripts. Grounded in 1,800 fine-grained expert ratings across the 60-item CTS-Psy instrument, this supervisory dataset is comparable in scale to prior benchmarks while reflecting full-length, real clinical encounters. Using prediction powered inference, we quantifying uncertainty on 288 unlabeled sessions by correcting for systematic biases. We demonstrate that fine-grained category decomposition paired with evidence-first reasoning achieves the highest accuracy and tightest human alignment, with a 9B model (Qwen3.5-9B) minimizing systematic evaluation bias. \textbf{(2) External validation:} We establish external construct validity on an independent RCT dataset of 102 transcripts, demonstrating that our approach reliably distinguishes CBT from non-CBT arms across core therapeutic skill dimensions. \textbf{(3) Feedback generation and evaluation:} Finally, we develop an automated pipeline that translates underlying fidelity assessments into structured, clinician-facing supervisor feedback. Through a structured user study with mental health experts, we demonstrate that the pipeline successfully synthesizes structured and actionable clinical suggestions.

\section{Related Work}

% Early work on automated fidelity assessment predicts CBT fidelity codes using transformer-based or neural models~\citep{creed2022enhancing, stanco2021lyssn}, scores motivational interviewing (MI) fidelity with LSTMs and support vector regression~\citep{flemotomos2022automated}, or pairs classical NLP features with machine learning classifiers to evaluate coach fidelity in guided digital CBT~\citep{zainal2025capitalizing}. A more recent line of work leverages LLMs to operationalize clinical rubrics directly, typically adapting the CTRS to score CBT adherence over multi-turn sessions. Within this paradigm, researchers have used these scores as reward signals for training therapy models~\citep{huang2026therapygym}, benchmarked LLM-generated therapy against expert-labeled sessions~\citep{chiu2024computational}, or compared model outputs against human clinicians for CBT~\citep{bedi2026assessing} and MI~\citep{teferra2026alignment}. Specifically, \citet{rollwage2026cognitive} show that enhanced LLMs can approach or exceed clinician-level CTRS competence at deployment scale.

Early work on automated psychotherapy fidelity assessment predicts CBT fidelity codes with transformer-based or neural models~\citep{creed2022enhancing, stanco2021lyssn}, scores motivational-interviewing fidelity with LSTMs and support vector regression~\citep{flemotomos2022automated}, or combines classical NLP features with machine learning to evaluate coach fidelity in guided digital CBT~\citep{zainal2025capitalizing}. A more recent line of work uses LLMs to operationalize clinical rubrics directly, typically adapting the CTRS to score CBT adherence over multi-turn sessions. Here, some use these scores as reward signals for training therapy models~\citep{huang2026therapygym}, others benchmark LLM-generated therapy against expert-labeled sessions~\citep{chiu2024computational} or compare model outputs against human clinicians in CBT~\citep{bedi2026assessing} and motivational interviewing~\citep{teferra2026alignment}. Specifically, \citet{rollwage2026cognitive} show enhanced LLMs can approach or exceed clinician-level CTRS competence at deployment scale.

% Early work on automated psychotherapy fidelity assessment predicts CBT fidelity codes with transformer-based or neural models~\citep{creed2022enhancing, stanco2021lyssn}, scores motivational-interviewing fidelity with LSTMs and support vector regression~\citep{flemotomos2022automated}, or combines classical NLP features with machine learning to evaluate coach fidelity in guided digital CBT~\citep{zainal2025capitalizing}. A more recent line of work uses LLMs to operationalize clinical rubrics directly, typically adapting the CTRS to score CBT adherence over multi-turn sessions: some use these scores as reward signals for training therapy models~\citep{huang2026therapygym}, others benchmark LLM-generated therapy against expert-labeled sessions~\citep{chiu2024computational} or compare model outputs against human clinicians in CBT~\citep{bedi2026assessing} and motivational interviewing~\citep{teferra2026alignment}, and \citet{rollwage2026cognitive} show enhanced LLMs can approach or exceed clinician-level CTRS competence at deployment scale.

Our work differs from this LLM-based line of work in three ways. First, these studies evaluate LLMs in the role of the therapist, scoring synthetic, role-played, or model-generated dialogue; we instead use LLMs as a therapist-facing assessment tool applied to authentic sessions delivered by human therapists to real patients, providing scalable fidelity feedback rather than benchmarking generated therapy. Second, we assess fidelity using the CTS-Psy rather than the more general CTRS, capturing psychosis-specific competencies that CTRS does not target. Third, prior work typically treats fidelity scoring as a single end-to-end judgment; we instead treat it as a task whose accuracy depends on inference structure, decomposing the rubric and comparing reasoning strategies, and we evaluate generalization beyond a single dataset.

\section{Data} 

\subsection{Study: Open Trial}

Our open trial examined the effectiveness of in-person therapy combined with mobile sensing for improving social functioning in individuals with SMI. Here we describe only the aspects relevant to therapy fidelity estimation. The intervention consisted of eight weekly, 60-minute individual therapy sessions delivered in person to 46 participants aged 18 to 65 over a 24-week period. Eligibility required a clinical diagnosis of schizophrenia, schizoaffective disorder, bipolar I disorder with psychotic features, or depression with psychosis \citep{nimh_mental_illness}, along with clinically significant social withdrawal and stable medication status; demographics, full inclusion, and exclusion criteria are listed in Appendix~\S\ref{apd:data}. 

% Our open trial examined the effectiveness of in-person therapy combined with mobile sensing for improving social functioning in individuals with SMI. Here we describe only the aspects relevant to therapy fidelity estimation. The intervention consisted of eight weekly, 60-minute individual therapy sessions delivered in person. We enrolled 46 participants aged 18 to 65 over a 24-week period. To be eligible, participants required: (1) a clinical diagnosis of schizophrenia, schizoaffective disorder, bipolar I disorder with psychotic features, or depression with psychosis \citep{nimh_mental_illness}; (2) moderate to high social withdrawal, indicated by a score of 2 or above on the asociality item of the Scale for the Assessment of Negative Symptoms (SANS) \citep{andreasen1981scale}; (3) a reading proficiency at or above the sixth-grade level on the Wide Range Achievement Test-4 Reading subtest \citep{wilkinson1993wide} (English proficiency); and (4) stable medication use, with no hospitalizations or medication changes in the preceding four months. Participants were excluded if they had received CBT in the past two years, showed significant disorganization on the Positive and Negative Syndrome Scale (PANSS) \citep{kay1987positive}, or had a diagnosis of alcohol or substance dependence within the last three months per DSM-5 criteria.
% The sample was 39\% female and 52\% non-White, with a mean age of 48.1 years (SD = 9.4). Diagnoses included schizophrenia (46\%), schizoaffective disorder (43\%), bipolar disorder with psychosis (9\%), and major depression with psychosis (2\%).

\textbf{Therapy.} Each participant received eight weekly, 60-minute in-person CBT-based therapy sessions delivered by Master's-level therapists, who guided patients in developing skills to support their social goals through cognitive restructuring, a structured method for identifying and challenging dysfunctional thoughts~\citep{burns1978cognitive, clark2011cognitive, clark2013cognitive}, and the 3Cs framework to reduce defeatist beliefs and social avoidance. A total of 331 therapy sessions were obtained, of which 13 were excluded due to incomplete or corrupted recordings, yielding a final sample of 318 sessions (45 participants).
% Each participant received eight weekly, 60-minute in-person therapy sessions employing CBT-based approaches to address their mental health needs. Sessions were conducted by Master's-level therapists who guided patients in developing skills to support their social goals. Central to this process was cognitive restructuring, a structured CBT method for identifying and challenging dysfunctional thoughts~\citep{burns1978cognitive, clark2011cognitive, clark2013cognitive}. By working toward individually defined recovery goals, patients used the 3Cs to reduce defeatist beliefs and social avoidance. A total of 331 therapy sessions were obtained, of which 13 were excluded due to incomplete or corrupted recordings, yielding a final sample of 318 sessions from 45 participants for analysis.

\textbf{CTS-Psy Scale}~\citep{haddock2001investigation}. The CTS-Psy assesses fidelity across two 5-category sections (60 items total per session; Appendix~\S\ref{appendix:cts_psy}): general skills (Table~\ref{tab:cts_general}; agenda, feedback, understanding, interpersonal effectiveness, collaboration) and specific skills (Table~\ref{tab:cts_specific}; guided discovery, key cognitions, choice of interventions, homework, social skills training). Items are rated on a non-sequential scale: 0 (inappropriately omitted), 1 (appropriately included), or 9 (appropriately omitted). We map both 1 and 9 directly to a binary compliance score of 1, as both represent "clinical non-errors" where the therapist acted appropriately. As the original instrument collapses 9 to 1 post-hoc, we sought to target this final compliance threshold directly to streamline inference. Note that, if a therapist establishes an agenda, item 2 of that category receives a 1, otherwise 0. 

Two expert supervisors rated 30 unique session transcripts sampled from the 318 recordings, with an overlapping 10-session subset evaluated by both to assess reliability (Cohen's $\kappa=0.71$, substantial agreement). Across these 30 full-length sessions evaluated on the 60-item instrument, our protocol yields 1,800 fine-grained expert judgments. This volume is comparable to prior work while extending evaluation to long-form, real clinical dialogues (Table~\ref{tab:prior_work_datasets}; Appendix~\S\ref{apd:data_density}). Notably, reliability is computed over 600 ratings (10 sessions), exceeding the sample size needed to detect $\kappa\ge0.70$~\citep{sim2005kappa}. For the final 30-session evaluation dataset, we randomly selected one supervisor's ratings for the 10 dual-rated sessions.

% On the 10 sessions, we obtained an overall Cohen's $\kappa = 0.68$ (moderate agreement). Two categories (understanding and interpersonal effectiveness) were excluded from this calculation because a high ceiling effect rendered the metric uninterpretable; there was only a single disagreement, and one rater provided completely invariant responses (all 1.0). This is unsurprising, as these foundational clinical skills are consistently demonstrated across sessions.

\subsection{Study: Randomized Controlled Trial (RCT)}
To validate our fidelity estimator, we use therapy sessions (N=102) from a live, independent randomized controlled trial (RCT). This trial applies the same inclusion criteria as our open trial but randomizes participants into a supportive contact condition (control; n=9) or a CBT condition (intervention; n=13). Per protocol, each participant is offered up to 8 sessions. A single therapist, distinct from those in the open trial, delivered both arms; condition differences therefore cannot be attributed to therapist identity. Because the trial is ongoing, participants have completed differing numbers of sessions, so the analyzed sample comprises of 102 sessions (57 CBT, 45 non-CBT). By design, control-arm sessions are expected to score lower than CBT-arm sessions on the CTS-Psy specific section. We therefore hypothesize that our fidelity estimator will mirror this pattern, yielding lower scores for the control arm and higher scores for the intervention arm. This constitutes a known-groups validity test, assessing whether the LLM can reliably distinguish CBT (N=57) from non-CBT (N=45) sessions using total fidelity scores alone. 

\subsection{Datasets}
We utilize three datasets derived from the two studies for our analysis: \textbf{(D1) Open Gold:} 30 expert-rated sessions providing gold-standard ground truth. This dataset is used to compute performance metrics and calibrate the LLM pipeline. \textbf{(D2) Open Unlabelled:} 288 unrated sessions, used to estimate the statistical uncertainty of the \textit{Open Gold} findings. \textbf{(D3) RCT:} 102 transcripts used for external validation to determine if our approach extends to an independent study population. Audio recording transcription, de-identification, and pre-processing details are provided in Appendix~\S\ref{apd:data}.

\section{Methods}
\subsection{Fidelity Estimation} \label{sec:fidelity_estimation}

Given a deidentified therapy transcript $T$, a set of $m$ CTS-Psy items $Q = \{q_1, q_2, \dots, q_m\}$, and a prompt template function $\mathcal{P}$, we frame fidelity estimation as a multi-label binary classification problem. The prompt function $\mathcal{P}$ maps the unstructured transcript and evaluation criteria into a prompt $I = \mathcal{P}(T, Q)$. We formalize the LLM as a parameterized mapping $F_\theta : \mathcal{I} \to \{0, 1\}^m$, where $\mathcal{I}$ denotes the prompt space. The model processes the prompt $I$ to produce a binary decision vector: $Y = [y_1, y_2, \dots, y_m]^T \in \{0, 1\}^m$. Here, $y_i = 1$ if the therapist adhered to criterion $q_i$ (either by successfully performing the clinical action or appropriately omitting it), and $y_i = 0$ otherwise. To evaluate how task complexity and cognitive load affect estimation accuracy across the 60-item scale, we analyze the level of decomposition and reasoning required for accurate fidelity assessment.

\subsubsection{Decomposition}
 This refers to the cardinality and thematic composition of the item subset $Q$ presented to the LLM during a single inference pass. We evaluate five distinct task decomposition configurations from a spectrum of low to high decomposition: \textbf{(1) Full} ($|Q| = 60$): The baseline configuration where all CTS-Psy items are evaluated simultaneously by a single LLM call without any task partitioning. \textbf{(2) Observability}: Items are partitioned based on their clinical observability into an Explicit group ($|Q| = 24$), which consists of highly manifest behaviors (agenda, feedback, homework, and social), and an Implicit group ($|Q| = 36$), which requires deeper latent inference (understanding, interpersonal effectiveness, collaboration, discovery, cognition, and intervention). \textbf{(3) Section}: Items are naturally divided by scale sections into the General clinical skills section ($|Q| = 30$) and the specific skills section ($|Q| = 30$), with each evaluated independently. \textbf{(4) Clinical}: Items are decomposed into three core clinical functional domains: Relationship ($|Q| = 18$; capturing understanding, interpersonal effectiveness, and collaboration), Structure ($|Q| = 18$; capturing agenda, feedback, and homework), and Delivery ($|Q| = 24$; capturing discovery, cognition, intervention, and social items). \textbf{(5) Category} ($|Q| = 6$): The finest level of granularity we evaluate, where each of the 10 distinct CTS-Psy categories are evaluated across 10 independent inference passes.

\subsubsection{Reasoning}
We investigate three distinct inference procedures that vary in their structural design and computational cost: \textbf{(1) Rationalization (Predict-then-Explain):} To evaluate the model's immediate scoring capabilities under a direct mapping strategy, we adopt a "predict-then-explain" paradigm~\citep{rajani2019explain}. The LLM first generates the binary score $y_i$ for an item $i$ and subsequently rationalizes its choices by producing supporting textual evidence. \textbf{(2) Evidential Estimation (Extract-Assess-Predict):} This approach implements a progressive, bottom-up chain of reasoning: the model first isolates relevant clinical text spans from the transcript $T$, assesses these extracted spans against the operational definitions of the target CTS-Psy items $Q$, and finally produces the binary fidelity score based on that synthesis. This is functionally an explain-then-predict strategy, the ordering standard CoT prompting typically uses. However, we constrain the "explain" step to evidence extraction rather than free-form reasoning, since we lack ground-truth clinician reasoning chains. This constraint also mirrors how human raters apply the CTS-Psy in practice: identifying evidence before judging it against the criterion. \textbf{(3) Self-Reflection (Critic-in-the-Loop):} This approach extends evidential estimation by introducing an iterative critique-and-revision loop~\citep{madaan2023self}. Upon generation of the initial evidential response, a secondary LLM instantiated to act as an expert clinical supervisor evaluates each rating across three strict dimensions: rubric adherence, evidence coverage, and rater confidence calibration. This critical evaluation is fed back to the primary generator as supervisor feedback, which then executes a final revision of the binary decisions and supporting evidence. Prompt templates are presented in Appendix~\S\ref{apd:prompt_templates}.

\subsection{Feedback Generation}
After fidelity estimation, we investigate whether the best-performing LLM can translate these ratings into feedback that is both actionable and therapist-facing. For each session, we use the fidelity decision vector $Y$ to identify the three specific competency categories (from guided discovery, key cognitions, choices of intervention, homework, and social skills training) with the lowest total score, and construct an evidence set $E$ from the 18 items spanning these categories, where each item's evidence tuple $(y_i, s_i, a_i)$ consists of its binary score, \texttt{relevant\_spans}, and \texttt{criterion\_assessment} from evidential fidelity estimation. A generation LLM is then prompted with $E$ to produce feedback consisting of one sentence of positive feedback grounded in the session, followed by one concrete, behaviorally specific suggestion per weak competency area. We compare two feedback conditions using the same LLM backbone: (1) Basic: direct feedback generated from $E$ alone, and (2) RAG: This pipeline retrieves passages relevant to $E$ from a manual vector store of the CTS-Psy therapy manual and prepends them to the prompt.

\section{Experiments}

\subsection{Models}

We evaluate our framework across three models spanning different architectures and capacities: Qwen3.5-9B~\citep{team2026qwen3}, Gemma4-E4B-it, and Gemma4-26B-A4B-it~\citep{google2026gemma4}. By testing models ranging from compact, edge-deployable sizes (4–9B) to larger configurations (26B), we can determine whether advanced decomposition strategies require emergent capabilities native to large parameter frontiers or if they remain effective at smaller scales. Implementations details are provided in Appendix~\S\ref{apd:implementation_details}.

\subsection{Tasks \& Metrics} 
\textbf{(T1) Fidelity Performance.} The Open Gold dataset contains ground-truth labels, allowing us to evaluate classification quality directly. We select balanced accuracy (BA) to ensure equal weighting across heavily and lightly represented classes, and Matthew’s correlation coefficient (MCC) to provide a rigorous, balanced assessment of all four confusion matrix quadrants. Further, we include 95\% confidence intervals (CI) computed using the bootstrap method (N=1000). 

\textbf{(T2) Fidelity Uncertainty.} To address the absence of ground-truth labels in the Open Unlabelled dataset, we utilize the Prediction-Powered Inference (PPI) framework~\citep{angelopoulos2023prediction}. As unverified LLM outputs introduce systematic biases, we leverage PPI to rectify the raw model baseline (${\mu}_{\text{unlabeled}}$) using the expert-annotated Open Gold dataset. The framework adjusts the raw baseline to output an unbiased population prevalence rate ($\hat{\mu}_{\text{PPI}}$) by computing a statistical rectifier, $\Delta = \hat{\mu}_{\text{PPI}} - {\mu}_{\text{unlabeled}}$, derived from the validation subset. Furthermore, PPI constructs valid, distribution-free CIs. Methodologically, a smaller absolute delta ($|\Delta| \rightarrow 0$) and a narrower CI indicate a superior, more consistent model whose clinical reasoning tightly aligns with human experts, informing our statistical uncertainty while maximizing downstream inferential power. 

\textbf{(T3) Known-Groups Validity Test.} We test whether our framework can distinguish the control (Non-CBT) and intervention (CBT) arms in the RCT dataset. For each of five CBT skills (guided discovery, key cognitions, choices of intervention, homework, and social skills training), we fit a linear mixed-effects model with arm as the fixed effect and participant as a random intercept, and correct for multiple comparisons across skills using a permutation-based maximum-statistic (max-T) procedure~\citep{westfall1993resampling} with 1,000 participant-level permutations, preserving each participant's repeated observations under the null. The corrected p-value for each skill is the proportion of permuted max-T statistics at least as extreme as the observed $|z|$, with continuity correction.
% $p = (1 + |\{|z|_{\text{null}} \geq |z|_{\text{observed}}\}|) / (1 + 1{,}000)$.

\textbf{(T4) Social Skills Administration Validity Test.} Because the therapy manual explicitly prescribes social skills training during weeks 7--8, these are expected to be higher in this period than in weeks 1--6. We test this with a linear mixed-effects model, with total social fidelity score as the outcome, time block (weeks 1--6 as reference) as a fixed effect, and a random intercept for participant to account for within-subject correlation; the time-block coefficient tests whether weeks 7--8 fidelity differs significantly from weeks 1--6.

\begin{table*}[htbp]
\centering
\footnotesize 
\setlength{\tabcolsep}{3pt} 

\begin{tabular}{l l ccc c ccc c ccc}
\toprule
\multirow{2}{*}{\textbf{Decomposition}} & \multirow{2}{*}{\textbf{Metric}} & \multicolumn{3}{c}{\textbf{Qwen3.5-9B}} & & \multicolumn{3}{c}{\textbf{Gemma4-E4B-it}} & & \multicolumn{3}{c}{\textbf{Gemma4-26B-A4B-it}} \\
\cmidrule(lr){3-5} \cmidrule(lr){7-9} \cmidrule(lr){11-13}
 & & \textbf{RZ} & \textbf{EV} & \textbf{RF} & & \textbf{RZ} & \textbf{EV} & \textbf{RF} & & \textbf{RZ} & \textbf{EV} & \textbf{RF} \\
\midrule

\multirow{2}{*}{\textbf{Category}} & BA  & \stat{0.67}{0.64}{0.69} & \stat{\underline{\textbf{0.72}}}{0.70}{0.74} & \stat{0.65}{0.62}{0.67} & & \stat{0.62}{0.60}{0.64} & \stat{0.61}{0.59}{0.63} & \stat{0.57}{0.56}{0.59} & & \stat{0.58}{0.56}{0.60} & \stat{0.67}{0.65}{0.69} & \stat{0.67}{0.64}{0.69} \\
                                   & MCC & \stat{0.38}{0.33}{0.42} & \stat{\underline{\textbf{0.45}}}{0.41}{0.49} & \stat{0.29}{0.24}{0.33} & & \stat{0.34}{0.29}{0.39} & \stat{0.35}{0.30}{0.39} & \stat{0.22}{0.17}{0.27} & & \stat{0.32}{0.29}{0.36} & \stat{0.43}{0.39}{0.48} & \stat{0.39}{0.34}{0.43} \\
\addlinespace[0.6em]

\multirow{2}{*}{\textbf{Clinical}} & BA  & \stat{0.64}{0.62}{0.66} & \stat{0.64}{0.61}{0.67} & \stat{0.61}{0.59}{0.65} & & \stat{0.64}{0.63}{0.67} & \stat{0.66}{0.64}{0.68} & \stat{0.65}{0.63}{0.67} & & \stat{0.56}{0.55}{0.57} & \stat{0.69}{0.67}{0.71} & \stat{0.70}{0.68}{0.72} \\
                                   & MCC & \stat{0.35}{0.30}{0.39} & \stat{0.31}{0.24}{0.37} & \stat{0.24}{0.20}{0.33} & & \stat{0.38}{0.33}{0.46} & \stat{0.39}{0.34}{0.44} & \stat{0.38}{0.32}{0.42} & & \stat{0.28}{0.23}{0.31} & \stat{0.46}{0.42}{0.50} & \stat{0.45}{0.40}{0.50} \\
\addlinespace[0.6em]

\multirow{2}{*}{\textbf{Section}}  & BA  & \stat{0.65}{0.63}{0.67} & \stat{0.65}{0.63}{0.67} & \stat{0.65}{0.62}{0.67} & & \stat{0.67}{0.65}{0.69} & \stat{\underline{\textbf{0.68}}}{0.66}{0.70} & \stat{0.67}{0.65}{0.69} & & \stat{0.53}{0.52}{0.54} & \stat{0.63}{0.61}{0.65} & \stat{0.68}{0.65}{0.70} \\
                                   & MCC & \stat{0.38}{0.33}{0.42} & \stat{0.33}{0.29}{0.39} & \stat{0.35}{0.30}{0.39} & & \stat{0.42}{0.38}{0.46} & \stat{0.42}{0.37}{0.46} & \stat{0.40}{0.36}{0.45} & & \stat{0.21}{0.17}{0.24} & \stat{0.36}{0.31}{0.41} & \stat{0.42}{0.37}{0.46} \\
\addlinespace[0.6em]

\multirow{2}{*}{\textbf{Obs}}      & BA  & \stat{0.65}{0.63}{0.67} & \stat{0.65}{0.63}{0.67} & \stat{0.62}{0.60}{0.65} & & \stat{0.64}{0.62}{0.66} & \stat{0.66}{0.64}{0.68} & \stat{0.62}{0.60}{0.64} & & \stat{0.52}{0.51}{0.53} & \stat{0.68}{0.66}{0.70} & \stat{0.68}{0.66}{0.71} \\
                                   & MCC & \stat{0.40}{0.35}{0.44} & \stat{0.34}{0.30}{0.39} & \stat{0.29}{0.24}{0.34} & & \stat{0.40}{0.35}{0.44} & \stat{0.42}{0.37}{0.46} & \stat{0.34}{0.30}{0.39} & & \stat{0.18}{0.13}{0.21} & \stat{0.48}{0.43}{0.52} & \stat{0.45}{0.40}{0.49} \\
\addlinespace[0.6em]

\multirow{2}{*}{\textbf{Full}}     & BA  & \stat{0.60}{0.58}{0.62} & \stat{0.61}{0.59}{0.64} & \stat{0.59}{0.57}{0.60} & & \stat{0.67}{0.65}{0.69} & \stat{0.65}{0.62}{0.67} & \stat{0.63}{0.61}{0.65} & & \stat{0.55}{0.54}{0.56} & \stat{0.69}{0.66}{0.70} & \stat{\underline{\textbf{0.71}}}{0.69}{0.73} \\
                                   & MCC & \stat{0.26}{0.21}{0.31} & \stat{0.27}{0.22}{0.31} & \stat{0.26}{0.20}{0.31} & & \stat{\underline{\textbf{0.44}}}{0.40}{0.48} & \stat{0.40}{0.35}{0.44} & \stat{0.37}{0.32}{0.42} & & \stat{0.26}{0.21}{0.29} & \stat{0.49}{0.45}{0.53} &\stat{\underline{\textbf{0.51}}}{0.47}{0.55} \\
\bottomrule
\end{tabular}
\caption{\textbf{Evaluation of reasoning and decomposition strategies for fidelity estimation (T1).} Performance metrics report Balanced Accuracy (BA) and Matthews Correlation Coefficient (MCC), with 95\% confidence intervals provided in brackets. Reasoning strategies: Rationalization (RZ), Evidential Estimation (EV), and Reflection (RF).}
\label{tab:overall_results}
\end{table*}

\textbf{(T5) Feedback Generation: Paired Evaluation.} We compare RAG and Basic feedback in a paired setting, using the model with the best fidelity performance to keep the LLM backbone consistent across fidelity assessment and feedback generation. Generated feedback is evaluated on two dimensions, Actionability and Clarity (exact definitions in Appendix~\S\ref{apd:user_study}). (T5.1) \textit{LLM-as-a-judge}: two LLMs from distinct families (Nemotron-3-30B, gpt-oss-20B) each rate a randomized pair of feedbacks (RAG vs. Basic) on both dimensions using a 3-point scale (-1: A preferred, 0: neutral, +1: B preferred). (T5.2) \textit{Expert user study}: three experts rate a stratified sample of 30 RAG/Basic feedback pairs (15 low-, 10 medium-, 5 high-similarity; Appendix~\S\ref{apd:feedback_generation_evaluation}). This three-member panel balances recruitment feasibility~\citep{ayers2023comparing} with optimal reliability design, where adding raters beyond three yields limited precision gains~\citep{shoukri2004sample}. We report raw agreement and Cohen's $\kappa$ for clinician–clinician, judge–judge, and clinician–judge pairs, and assess RAG vs. Basic differences with linear mixed-effects models (preference score as outcome, fixed intercept, random intercept grouped by question).

\textbf{(T6) Feedback Generation: Absolute Evaluation.} T5 compares feedback pairs but does not capture absolute quality. We therefore ask three mental health experts to rate 30 generated feedback samples (15 RAG, 15 Basic) on a 1--5 Likert scale across three dimensions: (1) Actionability, (2) Clarity, and (3) Usefulness. Exact rating prompts are provided in Appendix~\S\ref{apd:user_study}.

\section{Results}

\subsection{Fidelity Estimation}

\begin{figure}[htbp]
    \centering
    \includegraphics[width=0.9\linewidth]{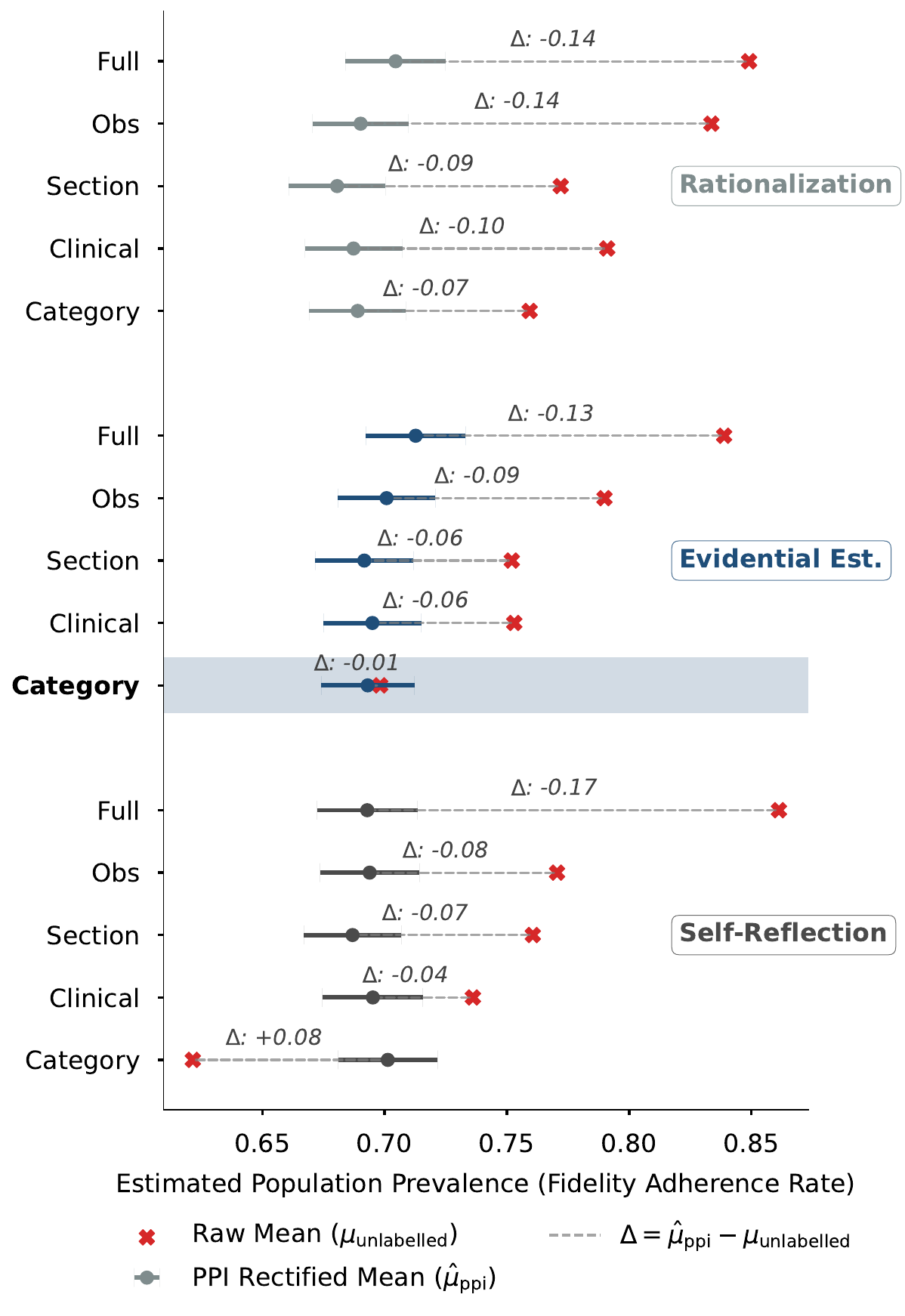}
    \caption{\textbf{Prediction Powered Inference of overall session fidelity prevalence using Qwen3.5-9B (T2).} Red crosses denote the biased, uncorrected raw population averages (${\mu}_{\text{unlabeled}}$) produced by each model configuration. The colored dots and corresponding horizontal brackets represent the unbiased PPI rectified means ($\hat{\mu}_{\text{PPI}}$) and their 95\% confidence intervals. Grey dashed lines illustrate the direction and magnitude of the statistical rectifier, defined as $\Delta = \hat{\mu}_{\text{PPI}} - {\mu}_{\text{unlabeled}}$.}
    \label{fig:ppi}
\end{figure}

\textbf{Across reasoning strategies and decomposition levels, evidential estimation with category-level decomposition on Qwen3.5-9B emerges as the single best configuration.} On the Open Gold set (Table~\ref{tab:overall_results}), this configuration achieves the highest balanced accuracy ($\text{BA} = 0.72$ [95\% CI: $0.70, 0.74$]) and MCC ($0.45$ [95\% CI: $0.41, 0.49$]), a substantial improvement over its own Full baseline ($\text{BA} = 0.61$, $\text{MCC} = 0.27$) and over competing Qwen decompositions and reasoning strategies (Table~\ref{tab:overall_results}). While Gemma4-26B-A4B-it yields a numerically higher peak MCC ($0.51$ [95\% CI: $0.47, 0.55$]), its confidence intervals overlap with EV-Category. Crucially, PPI analysis (Figure~\ref{fig:ppi}) reveals better alignment for Qwen3.5-9B EV-Category, minimizing systematic bias to near zero ($\Delta=-0.01$), whereas Gemma4 retains a negative bias at every configuration (Figures~\ref{fig:ppi_gemma4} \&~\ref{fig:ppi_gemma_4_26b}). Qwen3.5-9B EV-Category delivers the best accuracy and tightest human alignment at roughly a third of the total parameter count of Gemma4-26B, making it the best overall configuration.

% \textbf{Across reasoning strategies and decomposition levels, evidential estimation with category-level decomposition on Qwen3.5-9B emerges as the single best configuration.} On the Open Gold set (Table~\ref{tab:overall_results}), this configuration achieves the highest balanced accuracy ($\text{BA} = 0.72$ [95\% CI: $0.70, 0.74$]) and MCC ($0.45$ [95\% CI: $0.41, 0.49$]). This marks a substantial improvement over its own Full baseline ($\text{BA} = 0.61$, $\text{MCC} = 0.27$) and competing Qwen variants, which plateau at a BA of upto $0.67$ and an MCC of upto $0.40$. This gain stems specifically from evidence-grounded granularity; Rationalization and Self-Reflection at the Category level lag behind at MCCs of $0.38$ and $0.29$, respectively. While Gemma4-26B-A4B-it yields a numerically higher peak MCC ($0.51$ [95\% CI: $0.47, 0.55$], $\text{BA} = 0.71$), its confidence intervals overlap with EV-Category. Crucially, PPI analysis (Figure~\ref{fig:ppi}) reveals superior calibration for Qwen3.5-9B EV-Category, minimizing systematic bias to near zero ($\Delta = 0.01$) and narrowing CI width, whereas Gemma4 retains a negative bias (Figures~\ref{fig:ppi_gemma4} \&~\ref{fig:ppi_gemma_4_26b}). Qwen3.5-9B EV-Category therefore delivers the best accuracy and the tightest human alignment at roughly a third of the parameter count of Gemma4-26B, making it the best overall configuration on both performance and efficiency grounds.

\begin{figure}[htbp]
    \centering
    \includegraphics[width=1\linewidth]{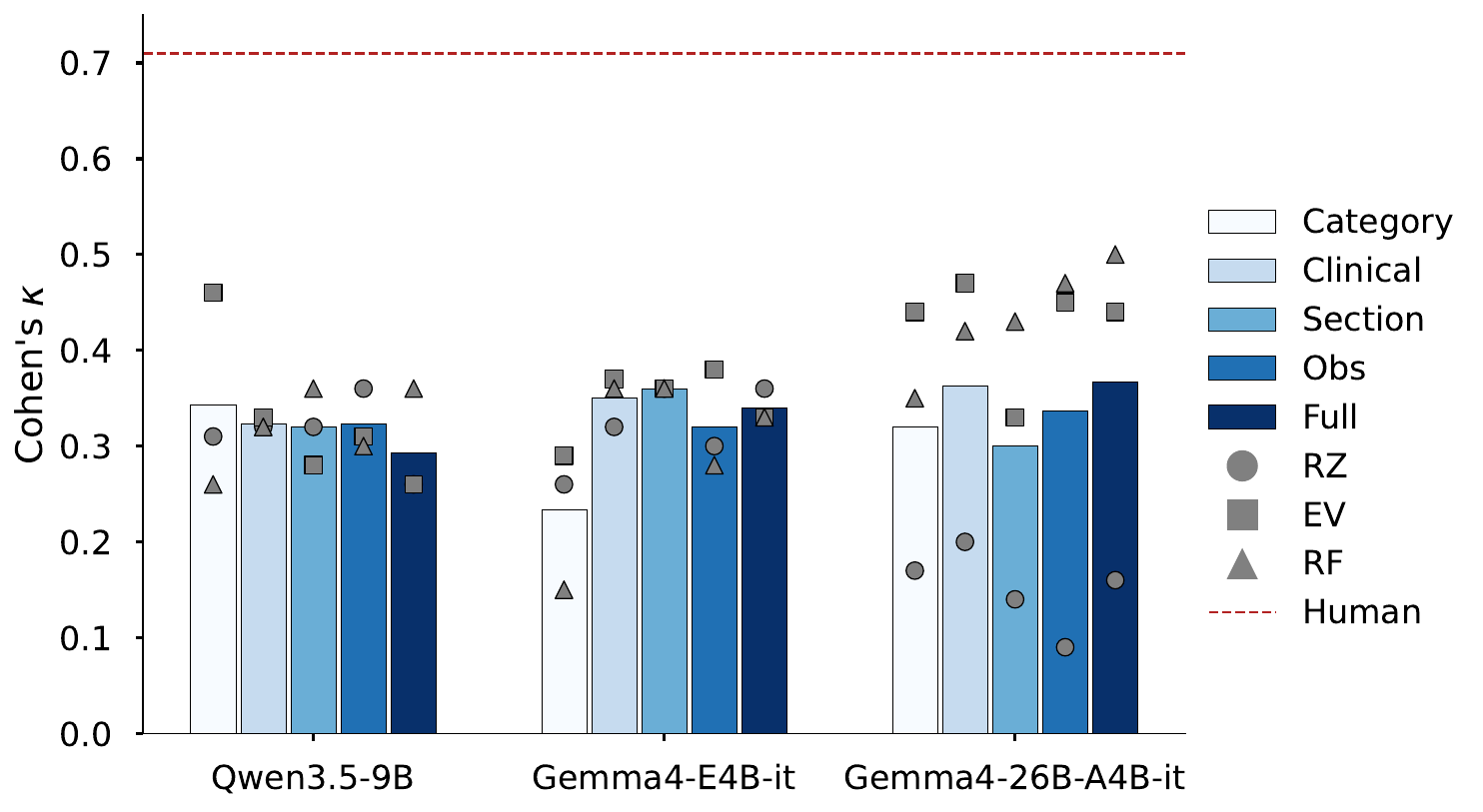}
    \caption{\textbf{Comparing LLM fidelity estimation performance vs. human performance (T1).} Performance reports Cohen's $\kappa$.}
    \label{fig:kappa_vs_human}
\end{figure}

\textbf{Our approach achieves moderate agreement in fidelity estimation across decomposition granularities, though a gap remains when compared to the human upper bound ($\kappa = 0.71$).} For Qwen3.5-9B, Evidential Estimation (EV) yields the strongest agreement at lower granularities (peaking at $\kappa = 0.46$ for Category), while Reflection (RF) is more effective at higher granularities (Figure~\ref{fig:kappa_vs_human}). Gemma4-E4B-it instead shows its best agreement under Rationalization at coarser decompositions. Interestingly, the larger Gemma4-26B-A4B-it struggles substantially with RZ reasoning but achieves the highest overall $\kappa$ in the study under RF and EV on Full decomposition ($\kappa = 0.50$ and $0.44$, respectively). Overall, EV and RF generally provide more robust fidelity alignment for larger architectures at higher granularities.

% \textbf{Our approach achieves moderate agreement in fidelity estimation across multiple decomposition granularities, though a gap remains when compared to the human upper bound ($\kappa = 0.71$).} For Qwen3.5-9B, Evidential Estimation (EV) yields the strongest agreement at lower granularities, peaking at $\kappa = 0.46$ for the Category level, though Reflection (RF) proves more effective at higher granularities, reaching $\kappa = 0.36$ on both Section and Full decompositions. Conversely, Gemma4-E4B-it shows more balanced performance across metrics, with rationalization performing particularly well on Section and Full decompositions ($\kappa = 0.36$). Interestingly, while the larger Gemma4-26B-A4B-it struggles significantly with RZ reasoning, dropping to $\kappa = 0.09$ for Obs-level data, it achieves the highest overall model performance in the study when utilizing Reflection (RF) and EV strategies on Full decompositions, reaching kappa scores of $0.50$ and $0.44$, respectively. Overall, EV and RF generally provide more robust fidelity alignment for larger architectures at higher granularities.

\textbf{The benefit of decomposition and reasoning is strongly model-dependent, and adequate model scale is necessary but not sufficient.} On Qwen3.5-9B, increasing decomposition granularity improves the predictive prevalence bias, culminating in $\Delta=-0.01$ for EV-Category (Figure~\ref{fig:ppi}), mirroring the accuracy gains from Full to Category level in Table~\ref{tab:overall_results}. Conversely, the smaller Gemma4-E4B-it exhibits large negative bias across all strategies and levels, with performance remaining flat regardless of decomposition (Figure~\ref{fig:ppi_gemma4}). Scale alone does not replicate Qwen's pattern: the larger Gemma4-26B-A4B-it yields its strongest performance under evidence-grounded reasoning ($\text{BA}$ up to $0.71$, $\text{MCC}$ up to $0.51$) while Rationalization collapses, and it maintains persistent negative bias at all decomposition depths (Figure~\ref{fig:ppi_gemma_4_26b}). Stronger reasoning does sequentially shrink this bias, from $\Delta \approx -0.24$–$-0.28$ under Rationalization to $-0.08$–$-0.12$ under Self-Reflection, but it never reaches the near-zero alignment of Qwen3.5-9B. Thus, low-bias fidelity estimation depends on a precise interaction between architecture and inference structure, not total parameter count. Given Qwen3.5-9B EV Category's balance of accuracy (T1) and predicted prevalence bias (T2), we adopt it for subsequent downstream analyses (T3--T6).

\begin{tcolorbox}[
    colback=gray!8,
    colframe=gray!60,
    boxrule=0.5pt,
    arc=1mm,
    left=4pt, right=4pt, top=3pt, bottom=3pt,
]
\textbf{Implication (T1--T2):} Qwen3.5-9B's strong accuracy and near-zero calibration bias show that locally deployable models can produce fidelity signal that is accurate and aligned. Given the remaining gap to human performance ($\kappa = 0.46$ vs. $0.71$), our approach is best positioned to augment expert review.
\end{tcolorbox}

\begin{figure}[htbp]
    \centering
    \includegraphics[width=1\linewidth]{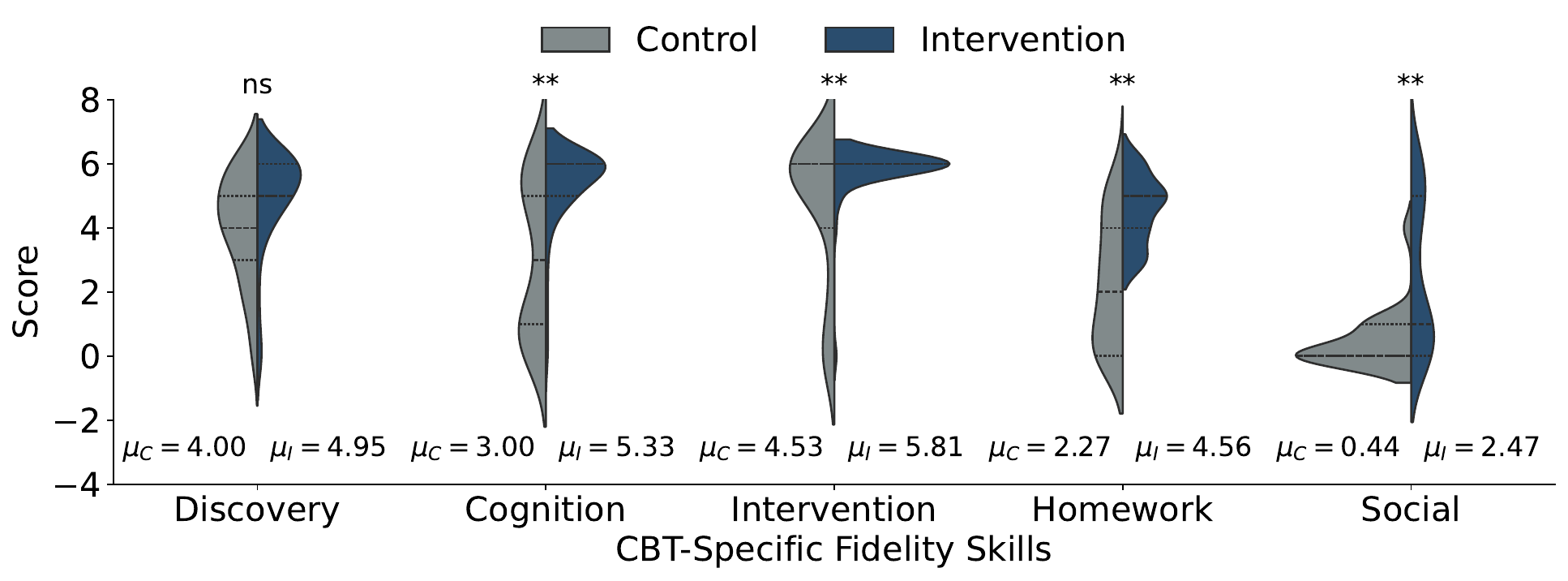}
    \caption{\textbf{Comparison of specific fidelity skill scores between control and intervention groups (T3).} The framework's ability to distinguish between the control group (Non-CBT, $\mu_C$) and the intervention group (CBT, $\mu_I$) is evaluated across five specific section skills. Statistical significant indicated as $^{**}:\ p < 0.01$, $^*:\ p < 0.05$.}
    \label{fig:known_groups_test}
    \vspace{-0.5cm}
\end{figure}

\textbf{Our approach establishes known-groups validity by producing significantly different fidelity scores between the intervention and control arms.} As shown in Figure~\ref{fig:known_groups_test}, the intervention group demonstrated higher mean scores across all five evaluated specific fidelity dimensions, with four differences statistically significant ($p < 0.01$): Key Cognitions, Choices of Intervention, Homework, and Social Skills Training. Guided Discovery was also higher in the intervention arm but not significantly so. These findings demonstrate that the category-level evidential decomposition distinguishes between CBT and non-CBT treatment arms by capturing distinct variations in fidelity skill execution, supporting its construct validity.

\textbf{Social fidelity scores are significantly higher in weeks 7--8, exactly when the manual mandates social skills training, reproducing the prescribed pattern.} A linear mixed-effects model estimated a mean social fidelity score of $1.30$ [95\% CI: $1.09, 1.51$] in weeks 1--6, rising by $1.92$ points [95\% CI: $1.54, 2.30$] in weeks 7--8 ($\beta = 1.92$, $z = 9.92$, $p < 0.001$), for a week 7--8 mean of $\approx 3.22$. This indicates that the inference strategy tracks protocol-driven variation in fidelity rather than spurious session-level features.

\begin{tcolorbox}[
    colback=gray!8,
    colframe=gray!60,
    boxrule=0.5pt,
    arc=1mm,
    left=4pt, right=4pt, top=3pt, bottom=3pt,
]
\textbf{Implication (T3--T4):} Passing two independent, pre-specified validity tests suggests the estimator's ratings reflect real clinical signal rather than model artifacts, supporting its use for pre-screening before allocating scarce expert review time.
\end{tcolorbox}

% \textbf{Social fidelity scores produced by Qwen3.5-9B with category-level evidential reasoning are significantly higher in weeks 7–8, exactly when the manual mandates social skills training, reproducing the prescribed pattern.} A linear mixed-effects model estimated a mean social fidelity score of $1.30$ [95\% CI: $1.09, 1.51$] in weeks 1–6 (intercept) and an increase of $1.92$ points [95\% CI: $1.54, 2.30$] in weeks 7–8 ($\beta = 1.92$, $\text{SE} = 0.19$, $z = 9.92$, $p < 0.001$), corresponding to a mean week 7–8 score of $\approx 3.22$. These results indicate that the inference strategy tracks protocol-driven variation in fidelity rather than spurious session-level features.

\subsection{Feedback Generation}

\textbf{Neither LLM judges nor clinicians preferred RAG over Basic feedback.} Across the full set, both judges showed no significant preference on either dimension (linear mixed-effects models, all $p > 0.05$). On the stratified 30-pair expert sample, clinicians showed no significant preference, with a non-significant lean toward the Basic baseline on actionability ($\beta = -0.21, p = 0.074$) and no difference on clarity ($\beta = -0.06, p = 0.59$); judges restricted to the same 30 pairs reproduced this pattern. Thus, retrieval did not improve feedback quality on either dimension. Inter-rater agreement declined from judges to clinicians and from actionability to clarity: the two judges reached moderate agreement (actionability $\kappa = 0.48$, clarity $\kappa = 0.45$; raw $0.70, 0.73$), while clinician agreement was substantially lower, slight on actionability (mean $\kappa = 0.15$) and near chance on clarity (mean $\kappa = -0.02$). This indicates that experts do not reliably agree on relative feedback quality, suggesting that choosing between comparable therapy feedback based on actionability and clarity is difficult even for trained clinicians. Consequently, we sought to determine whether this reflects two comparably strong approaches or two comparably weak ones.

\begin{figure}[htbp]
    \centering
    \includegraphics[width=\linewidth]{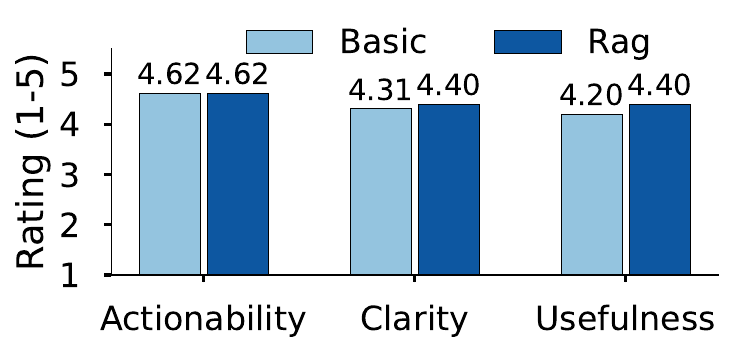}
    \caption{\textbf{Absolute quality evaluation of generated feedback (T6}). Mental health experts rated feedback on a 1–5 Likert scale across actionability, clarity, and usefulness.}
    \label{fig:ratings_absolute}
    \vspace{-0.5cm}
\end{figure}

\textbf{Absolute evaluations by mental health experts reveal that both configurations achieve high quality across all targeted dimensions}. This suggests that the lack of relative preference stems from a high performance ceiling rather than uniform mediocrity. Specifically, actionability was rated identically high for both the basic baseline and the RAG-grounded configuration, with both scoring 4.62 out of 5. The RAG-grounded feedback achieved minor numerical gains in clarity (4.40 vs. 4.31) and clinical usefulness (4.40 vs. 4.20). These results show that both pipelines produce high-quality feedback across all dimensions. Because the Basic feedback is sufficiently strong, experts found it difficult to establish a clear head-to-head preference.

\begin{tcolorbox}[
    colback=gray!8,
    colframe=gray!60,
    boxrule=0.5pt,
    arc=1mm,
    left=4pt, right=4pt, top=3pt, bottom=3pt,
]
\textbf{Implication (T5--T6):} Since retrieval adds no measurable benefit, a simpler pipeline may be preferable for deployment — reducing latency and infrastructure complexity without sacrificing clinical feedback quality.
\end{tcolorbox}

\section{Conclusion}

Our work presents an LLM application that systematically evaluates whether open-weight LLMs can automate psychotherapy fidelity assessment and generate high-quality supervisory feedback from real clinical transcripts. Clinical reasoning performance depends less on parameter scale than on the interplay between architecture and task decomposition. Under category-level decomposition and evidence-first reasoning, locally deployable models minimize systematic bias and support construct validity, while running locally to protect patient privacy. Our expert user study shows the pipeline produces behaviorally specific, actionable feedback closely matching expert judgment. In summary, our work offers a scalable application for bridging the mental health supervision gap and enhancing clinician training.

\section*{Limitations}
Several limitations are worth noting. First, our analysis relies on de-identified text transcripts rather than audio, since recordings were not retained; this omits paralinguistic cues (tone, pacing, affect) relevant to a few CTS-Psy items (e.g., empathetic tone under Understanding), which may partly explain the remaining LLM--human gap. Second, de-identification occasionally removes contextual details (e.g., names, places) relevant to fidelity judgments, an acceptable tradeoff given the population's sensitivity but one that may modestly affect transcript completeness. Third, our gold-standard set comprises 30 expert-rated sessions (10 doubly rated). While adequate for clinical evaluation studies of this scope (Appendix~\S\ref{apd:data_density}), larger samples would improve estimate precision. Given that optimal configurations are selected using this cohort, conclusions should be interpreted holistically alongside confidence intervals, bias assessments, and external validity. Fourth, our study population is drawn from two related trials targeting social functioning in SMI using a specific CBT-based protocol. Although we examined widely used therapeutic techniques such as agenda setting and homework review, future work should evaluate whether these findings generalize to other therapeutic modalities (e.g., motivational interviewing) and broader patient populations. Relatedly, while our external validation demonstrates construct validity, the unannotated RCT cohort precludes evaluating ground-truth accuracy. Future work should therefore benchmark classification performance on fully labeled external cohorts. Finally, we evaluate three open-weight models, chosen for their suitability to sensitive clinical data; results may differ for proprietary frontier models, and our findings suggest the benefits of decomposition and reasoning depend heavily on the underlying model.

\section*{Ethics Statement}

\textbf{Study.} Both the open trial and the independent RCT underlying this study were reviewed and approved by the relevant Institutional Review Board(s) prior to participant enrollment. All data collection, including audio recording of therapy sessions, was conducted under this approval, and secondary use of session transcripts for the fidelity-assessment analyses described in this paper falls within the scope of that approved protocol. Participants in both the open trial and RCT provided informed consent as part of the IRB-approved parent studies prior to any therapy session being recorded. Expert raters and clinicians involved in fidelity annotation and feedback evaluation were compensated for their time in accordance with institutional guidelines.

\textbf{Privacy and de-identification.} All therapy sessions were audio-recorded with participant consent under the parent clinical trials, transcribed locally using an open-weight speech recognition model, and subsequently de-identified using Presidio prior to any analysis. Only de-identified transcripts were retained for downstream processing; raw audio was not used in model inference. All model inference for the open-were performed locally rather than through third-party APIs, minimizing exposure of sensitive clinical content to external services.

\textbf{Intended use.} The proposed pipeline is designed to augment human clinical supervision, not replace it. Automated fidelity ratings and feedback reduce the burden on expert supervisors. They also provide timely, scalable feedback to community clinicians. Fidelity assessments are predictive and subject to error, which can lead to misdirected feedback. As such, this system should not be used to make autonomous decisions about clinician competence, employment, or patient care. The outputs from this system should be treated as a supervisory aid. They must always be subject to human review, especially before being used in any high-stakes evaluative context like clinician certification or performance reviews. We caution against deploying automated tools without human oversight. This is particularly important for populations with SMI. In those cases, miscalibrated ratings could discourage effective practices, or inaccurately flag appropriate care as non-adherent.

\textbf{Reproducibility.} We will publicly release the code alongside synthetic dummy transcripts mirroring our clinical schema upon publication. The underlying therapy transcripts contain sensitive clinical data and cannot be publicly released. 
% We will instead release our full pipeline code, including prompts, decomposition configurations, reasoning strategies, and evaluation scripts. We will also release synthetic, non-clinical example transcripts that mimic the structure and format of real sessions. This supports reproducibility without exposing patient data. 
To support reproducibility, we report full model configurations and inference settings in Appendix~\S\ref{apd:implementation_details}. We provide the complete CTS-Psy item set used for fidelity scoring (Appendix~\S\ref{appendix:cts_psy}). We describe our transcript preprocessing and de-identification pipeline (Appendix~\S\ref{apd:data}). We also include the exact prompt templates used for each reasoning strategy and feedback generation stage (Appendix~\S\ref{apd:prompt_templates}).

% Bibliography entries for the entire Anthology, followed by custom entries
%\bibliography{anthology,custom}
% Custom bibliography entries only
\bibliography{custom}

\appendix

\section{Use of AI Assistants}
\label{appendix:ai-use}

We used LLM-based assistants (Claude and Gemini) for writing assistance, including grammar correction, phrasing suggestions, and editing of draft text. We also used them for code assistance, specifically to help optimize multiprocessing and parallelization of our LLM inference calls (via vLLM). These assistants were run separately with no access to raw experimental data, transcripts, or session recordings; all model inference involving clinical data was performed locally as described in the Ethics Statement. All research ideas, experimental design, scientific claims, and the final text were authored and verified by the human authors.

\section{Cognitive Therapy Scale for Psychosis (CTS-Psy)}
\label{appendix:cts_psy}

The CTS-Psy is organized into two sections, each  comprising five categories of six items, for 60 items in total. Table~\ref{tab:cts_general} lists the General Items section, which assesses core therapeutic process  skills applicable across session types: agenda setting, feedback, understanding, interpersonal effectiveness, and collaboration. Table~\ref{tab:cts_specific} lists the Specific Items section, which assesses specific technique execution: guided discovery, key cognitions, choice of intervention, homework, and social skills training. We exclude the overall 
score category (item j) from our analysis, as it constitutes a holistic judgment rather than a discrete, independently verifiable criterion.

\clearpage
\onecolumn

\begin{longtable}{l c p{10cm}}
\toprule
\textbf{Category} & \textbf{Item} & \textbf{Description} \\ \midrule
\endfirsthead
\toprule
\textbf{Category} & \textbf{Item} & \textbf{Description} \\ \midrule
\endhead
\midrule \multicolumn{3}{r}{{Continued on next page}} \\ \bottomrule
\endfoot
\bottomrule
\caption{\textbf{CTS-Psy Section I:} General Items} \label{tab:cts_general} \\
\endlastfoot

\textbf{a) Agenda} & 1 & Therapist noted client's current emotional state regarding agenda setting. \\
 & 2 & Therapist and client established agenda for session. \\
 & 3 & Priorities for agenda items were established. \\
 & 4 & Agenda was appropriate for time allotment (neither too ambitious nor too limited). \\
 & 5 & Client discussed salient events or problems occurring since last session. \\
 & 6 & The agenda was adhered to during the session. \\ \midrule

\textbf{b) Feedback} & 1 & Therapist asked for feedback concerning previous session. \\
 & 2 & Therapist asked for feedback and reactions to present session. \\
 & 3 & Therapist asked specifically for any negative reactions to therapist, content, or formulation. \\
 & 4 & Therapist attempted to respond to client's feedback. \\
 & 5 & Therapist checked that client understood the therapist's role and session limitations. \\
 & 6 & Therapist summarized and asked client to fine-tune or correct perspectives. \\ \midrule

\textbf{c) Understanding} & 1 & Therapist conveys understanding by rephrasing or summarizing. \\
 & 2 & Therapist shows sensitivity by reflecting back feelings as well as ideas. \\
 & 3 & Therapist's tone of voice was empathetic. \\
 & 4 & Therapist acknowledged client's viewpoint as valid and important. \\
 & 5 & Therapist did not negate client's point of view. \\
 & 6 & Where differences occurred, they were acknowledged and mutually respected. \\ \midrule

\textbf{d) Interpersonal} & 1 & Therapist seemed open rather than defensive. \\
\textbf{Effectiveness} & 2 & Content communicated warmth, concern, and caring. \\
 & 3 & Therapist did not criticize, disapprove, or ridicule client behavior or point of view. \\
 & 4 & Therapist responded to, or displayed, humor when appropriate. \\
 & 5 & Therapist made clear statements without frequent hesitations or rephrasings. \\
 & 6 & Therapist was in control of the session, shifting between listening and leading. \\ \midrule

\textbf{e) Collaboration} & 1 & Therapist asked client for suggestions and offered choices when feasible. \\
 & 2 & Therapist ensured client's suggestions and choices were acknowledged. \\
 & 3 & Therapist explained rationale for intervention. \\
 & 4 & Flow of verbal interchange was smooth, with a balance of listening and talking. \\
 & 5 & Collaboration was maintained even when therapist was in an educational role. \\
 & 6 & Discussion was at a language level easily understandable by the client. \\
\end{longtable}

\begin{longtable}{l c p{10cm}}
\toprule
\textbf{Category} & \textbf{Item} & \textbf{Description} \\ \midrule
\endfirsthead
\toprule
\textbf{Category} & \textbf{Item} & \textbf{Description} \\ \midrule
\endhead
\midrule \multicolumn{3}{r}{{Continued on next page}} \\ \bottomrule
\endfoot
\bottomrule
\caption{\textbf{CTS-Psy Section II:} Specific Items. Note that we do not use the overall score category j).} \label{tab:cts_specific} \\
\endlastfoot

\textbf{f) Guided} & 1 & Used questions to determine meaning client attached to an event. \\
\textbf{Discovery} & 2 & Used questions to show incongruities without demeaning the person. \\
 & 3 & Used questions to help client explore various facets of a problem. \\
 & 4 & Used questions to examine client's arbitrary conclusions or assumptions. \\
 & 5 & Used questions to elicit alternative ways of solving a problem. \\
 & 6 & Used questions to consider alternative explanations. \\ \midrule

\textbf{g) Key Cognitions} & 1 & Elicited specific thoughts, assumptions, images, memories, or beliefs. \\
 & 2 & Cognitions elicited were those involved in key problems. \\
 & 3 & Discussed phenomenological characteristics (frequency, duration, etc.). \\
 & 4 & Discussed relationship with client's key problems. \\
 & 5 & Discussed the link between cognition and affect. \\
 & 6 & Discussions took place in an atmosphere of collaboration. \\ \midrule

\textbf{h) Choices of} & 1 & Selected cognitive-behavioral techniques of intervention. \\
\textbf{Intervention} & 2 & Overall strategy was specifically related to client's problems. \\
 & 3 & Each individual technique was relevant to a key problem. \\
 & 4 & Strategies used were directly related to formulation. \\
 & 5 & Techniques had demonstrable potential for change based on evidence. \\
 & 6 & Therapist sought feedback regarding the strategy for change. \\ \midrule

\textbf{i) Homework} & 1 & Therapist explicitly reviewed previous week's homework. \\
 & 2 & Therapist summarized conclusions or progress made from homework. \\
 & 3 & Appropriate homework was assigned. \\
 & 4 & Therapist explained rationale for homework assignment. \\
 & 5 & Homework was specific and details were clearly explained. \\
 & 6 & Therapist asked if client anticipated any problems in carrying out homework. \\ \midrule

\textbf{k) Social Skills} & 1 & Established rationale for skill by linking to client goals. \\
\textbf{Training} & 2 & Discussed and modeled steps of the skill in a role play. \\
 & 3 & Engaged client in a role play of the skill. \\
 & 4 & Provided behaviorally specific positive feedback and reinforcement. \\
 & 5 & Provided behaviorally specific suggestions for improvement. \\
 & 6 & Used a shaping approach by reinforcing small steps in repeated role plays. \\
 
\end{longtable}

\clearpage
\twocolumn
\section{Implementation Details} \label{apd:implementation_details}

All open-weight models were served locally using vLLM~\citep{kwon2023efficient} and queried via LangChain~\footnote{\url{https://www.langchain.com/}} on 4$\times$ NVIDIA RTX A4500 GPUs. 

\textbf{Qwen3.5-9B}~\citep{team2026qwen3} is a hybrid reasoning dense model. We use the following parameters in our setup: temperature 1.0, top-p 0.95, top-k 20, presence penalty 1.5 to encourage extended deliberation before producing structured outputs. In reflection runs, a separate instance with thinking disabled served as the critique model.

\textbf{Gemma4-E4B-it}~\citep{google2026gemma4} is a 4B parameter Mixture-of-Experts instruction-tuned model. We use the following parameters in our setup: temperature 1.0, top-p 0.95, top-k 64, presence penalty 1.5. In reflection runs, the critique model's output was capped at 512 tokens.

\textbf{Gemma4-26B-A4B-it}~\citep{google2026gemma4} is a 26-billion-parameter Mixture-of-Experts instruction-tuned model sharing the same inference parameters as Gemma 4B. Due to its size, the model was served across 4 GPUs with tensor parallelism, bfloat16 precision, and a 131k-token context window.

\textbf{gpt-oss-20b}~\cite{agarwal2025gpt} is a 20-billion-parameter open-weight reasoning model, used exclusively as an LLM-as-a-judge. We serve it locally with deterministic decoding (temperature 0.0, top-p 1.0) to minimize scoring variance across paired feedback comparisons.

\textbf{Nemotron-3-Nano-30B-A3B}~\citep{blakeman2025nemotron} is a 30-billion-parameter Mixture-of-Experts reasoning model, used only as an LLM-as-a-judge and served with deterministic decoding settings (temperature 0.0, top-p 1.0).

% Because both judge models interleave free-form reasoning with their final answer and do not reliably comply with schema-constrained decoding, we do not enforce structured output at generation time for either. Instead, the model reasons freely, and we parse the last well-formed JSON object appearing after any reasoning block (delimited by \texttt{<think>}...\texttt{</think>} tags where present) to recover the judge score for each dimension.

\section{Study \& Data} \label{apd:data}

\subsection{Open Trail: Eligibility Criteria} \label{apd:eligibility_criteria}

Participants were required to meet the following criteria: (1) a clinical diagnosis of schizophrenia, schizoaffective disorder, bipolar I disorder with psychotic features, or depression with psychosis \citep{nimh_mental_illness}; (2) moderate to high social withdrawal, indicated by a score of 2 or above on the asociality item of the Scale for the Assessment of Negative Symptoms (SANS) \citep{andreasen1981scale}; (3) a reading proficiency at or above the sixth-grade level on the Wide Range Achievement Test-4 Reading subtest \citep{wilkinson1993wide}; and (4) stable medication use, with no hospitalizations or medication changes in the preceding four months. Participants were excluded if they had received CBT in the past two years, showed significant disorganization on the Positive and Negative Syndrome Scale (PANSS) \citep{kay1987positive}, or had a diagnosis of alcohol or substance dependence within the last three months per DSM-5 criteria.

Overall, our study was 39\% female and 52\% non-White, with a mean age of 48.1 years (SD = 9.4). Diagnoses included schizophrenia (46\%), schizoaffective disorder (43\%), bipolar disorder with psychosis (9\%), and major depression with psychosis (2\%).

\subsection{Evaluation Comparison to Prior Work} \label{apd:data_density}

As shown in Table~\ref{tab:prior_work_datasets}, our evaluation expands both the scale and annotation granularity relative to existing benchmarks. While prior datasets rely on brief exchanges such as Cactus (16.6 turns) and TherapyGym (10 turns), our protocol evaluates full-length, authentic clinical interactions averaging approximately 180 turns per dialogue, representing an increase in dialogue length of more than 10$\times$. In total, our evaluation encompasses 420 complete sessions, spanning 30 gold-standard sessions, 288 unlabelled sessions for uncertainty quantification, and 102 sessions for external randomized controlled trial (RCT) validation. Across roughly 5,400 annotated authentic turns, our methodology provides 1,800 fine-grained expert judgments. This corresponds to an evaluation density of 60.0 rating items per dialogue, offering over 5$\times$ the granular supervision of TherapyGym (11.0 items) and substantially exceeding corpora like Cactus, which contain no itemized per-dialogue ratings. Grounding our evaluation in authentic clinical transcripts preserves critical ecological validity.

\begin{table*}[]
\centering
\begin{tabular}{@{}lllll@{}}
\toprule
\textbf{Study} & \textbf{\#Dialogues}  &\textbf{\#Turns (Form)}  &\textbf{Type}  &\textbf{\#Ratings}  \\ \midrule
Cactus~\citep{lee2024cactus} &\textbf{31577}  &16.6 (Short)  &Synthetic  & 0 (unrated)  \\ 
TherapyGym~\citep{huang2026therapygym} &116 &10 (Short) &Synthetic & 1276  \\
Ours &30 (420 full) &\textbf{$\sim$180$\pm$81 (Long)} &\textbf{Real} &\textbf{1800}  \\
\bottomrule
\end{tabular}
\caption{\textbf{Comparison of prior evaluations and data for dialogue-based therapy.}}
\label{tab:prior_work_datasets}
\end{table*}

\subsection{Pre-processing}
Since therapy audio recordings constitute personally identifiable information, we prioritized participant privacy throughout the data processing pipeline. We transcribed all audio files using a locally deployed Whisper-large-v3-turbo~\citep{radford2022whisper} model and subsequently de-identified them using Presidio~\citep{MicrosoftPresidio}, retaining only the de-identified transcripts for all downstream analyses. We then applied a standardized cleaning and normalization pipeline to ensure transcript quality. Specifically, we removed non-ASCII characters (e.g., Korean or Chinese text) and bracketed annotations such as stage directions or speaker labels, and merged consecutive turns from the same speaker into a single continuous block. We further pruned low-value filler content by dropping short therapist backchannels (e.g., "Okay.", "Yeah.") and removing common verbal fillers (e.g., um, uh, erm). Finally, we normalized the transcripts by collapsing redundant whitespace and correcting spacing irregularities.

\section{Additional Results}
\subsection{Fidelity Performance (T1)} \label{apd:fidelity_performance}
To evaluate the agreement between the automated pipelines and human expert judgments (T1), we present a comprehensive comparison of Cohen's $\kappa$ statistics across all architectural and decomposition configurations in Table~\ref{tab:fidelity_performance_kappa}. These results demonstrate that while a performance gap remains when compared to the human upper bound ($\kappa = 0.71$), utilizing evidential estimation (EV) or self-reflection (RF) strategies generally provides a more robust alignment than standard rationalization, particularly at granular decomposition levels.

\begin{table*}[ht] 
\centering
\setlength{\tabcolsep}{3pt} 

\begin{tabular}{l ccc c ccc c ccc}
\toprule
\textbf{Decomposition} & \multicolumn{3}{c}{\textbf{Qwen3.5-9B}} & & \multicolumn{3}{c}{\textbf{Gemma4-E4B-it}} & & \multicolumn{3}{c}{\textbf{Gemma4-26B-A4B-it}} \\
\cmidrule(lr){2-4} \cmidrule(lr){6-8} \cmidrule(lr){10-12}
 & \textbf{RZ} & \textbf{EV} & \textbf{RF} & & \textbf{RZ} & \textbf{EV} & \textbf{RF} & & \textbf{RZ} & \textbf{EV} & \textbf{RF} \\
\midrule

\textbf{Category} &0.31  &0.46  &0.26 & &0.26 &0.29  &0.15 &  &0.17  &0.44 &0.35   \\
\addlinespace[0.6em]

\textbf{Clinical} &0.32  &0.33  &0.32 & &0.32 &0.37  &0.36 & &0.20  &0.47 &0.42    \\
\addlinespace[0.6em]

\textbf{Section}  &0.32  &0.28  &0.36 & &0.36 &0.36  &0.36 & &0.14  &0.33 &0.43     \\
\addlinespace[0.6em]

\textbf{Obs}      &0.36  &0.31  &0.30 & &0.30 &0.38  &0.28 & &0.09  &0.45 &0.47     \\
\addlinespace[0.6em]

\textbf{Full}     &0.26  &0.26  &0.36  & &0.36 &0.33  &0.33 & &0.16  &0.44 &0.50   \\
\midrule
\textbf{Human} (Upper Bound) & \multicolumn{11}{c}{0.71} \\
\bottomrule
\end{tabular}
\caption{\textbf{Comparing LLM fidelity estimation performance vs. human performance (T1).} Performance reports Cohen's $\kappa$. Human performance upper bound is $\kappa = 0.71$. Reasoning strategies evaluated include Rationalization (RZ), Evidential Estimation (EV), and Reflection (RF).}
\label{tab:fidelity_performance_kappa}
\end{table*}

\subsection{Repeated-Run Stability}
LLM parameters are chosen based on recommendations. To evaluate the stability of stochastic decoding, we run the Qwen3.5-9B EV reasoning with category decomposition three times. We observe via majority voting that 90.1\% of all 19,080 ($318 \times 60$) labels remain unchanged. Across runs, balanced accuracy ($0.70$, $0.71$, $0.72$) and MCC ($0.41$, $0.42$, $0.45$) fall within our reported 95\% CIs ($[0.70, 0.74]$; $[0.41, 0.49]$).

\subsection{Fidelity Performance Across Categories (T1)}

Across categories, performance varied in both BA and MCC. The model achieved its highest performance on Social Skills (BA = 0.87 [0.80, 0.92]; MCC = 0.57 [0.43, 0.69]), followed by strong classification on Cognitions (BA = 0.67; MCC = 0.40), Agenda (BA = 0.68; MCC = 0.39), and Homework (BA = 0.71; MCC = 0.36). Conversely, performance dropped considerably on categories such as Discovery (BA = 0.55; MCC = 0.08), Feedback (BA = 0.54; MCC = 0.08), and Collaboration (BA = 0.53; MCC = 0.11). Notably, while Intervention and Interpersonal reached moderate to high balanced accuracy (0.71 and 0.78, respectively), their lower MCC scores (0.11 and 0.22) reflect substantial class imbalance and high variance across evaluations.

\begin{table}[]
\centering
\begin{tabular}{@{}lll@{}}
\toprule
\textbf{Category} & \textbf{BA} & \textbf{MCC} \\ \midrule
Agenda &\stat{0.68}{0.61}{0.75} &\stat{0.39}{0.24}{0.54} \\
Feedback &\stat{0.54}{0.47}{0.62} &\stat{0.08}{-0.06}{0.23} \\
Understanding &\stat{0.63}{0.48}{0.78} &\stat{0.26}{-0.04}{0.54} \\
Interpersonal &\stat{0.78}{0.44}{0.96} &\stat{0.22}{-0.04}{0.44} \\
Collaboration &\stat{0.53}{0.48}{0.59} &\stat{0.11}{-0.06}{0.28} \\
Discovery &\stat{0.55}{0.46}{0.63} &\stat{0.08}{-0.06}{0.21} \\
Cognitions &\stat{0.67}{0.59}{0.76} &\stat{0.40}{0.21}{0.58} \\
Intervention &\stat{0.71}{0.63}{0.80} &\stat{0.11}{-0.04}{0.31} \\
Homework &\stat{0.71}{0.63}{0.80} &\stat{0.36}{0.21}{0.50} \\
Social Skills &\stat{0.87}{0.80}{0.92} &\stat{0.57}{0.43}{0.69} \\
\bottomrule
\end{tabular}
\caption{\textbf{Qwen3.5-9B with evidential reasoning and category level decomposition performance across categories.}}
\label{tab:fidelity_across_categories}
\end{table}

\subsection{Fidelity Uncertainty (T2)} \label{apd:fidelity_uncertainity}

\begin{figure}[ht]
    \centering
    \includegraphics[width=1\linewidth]{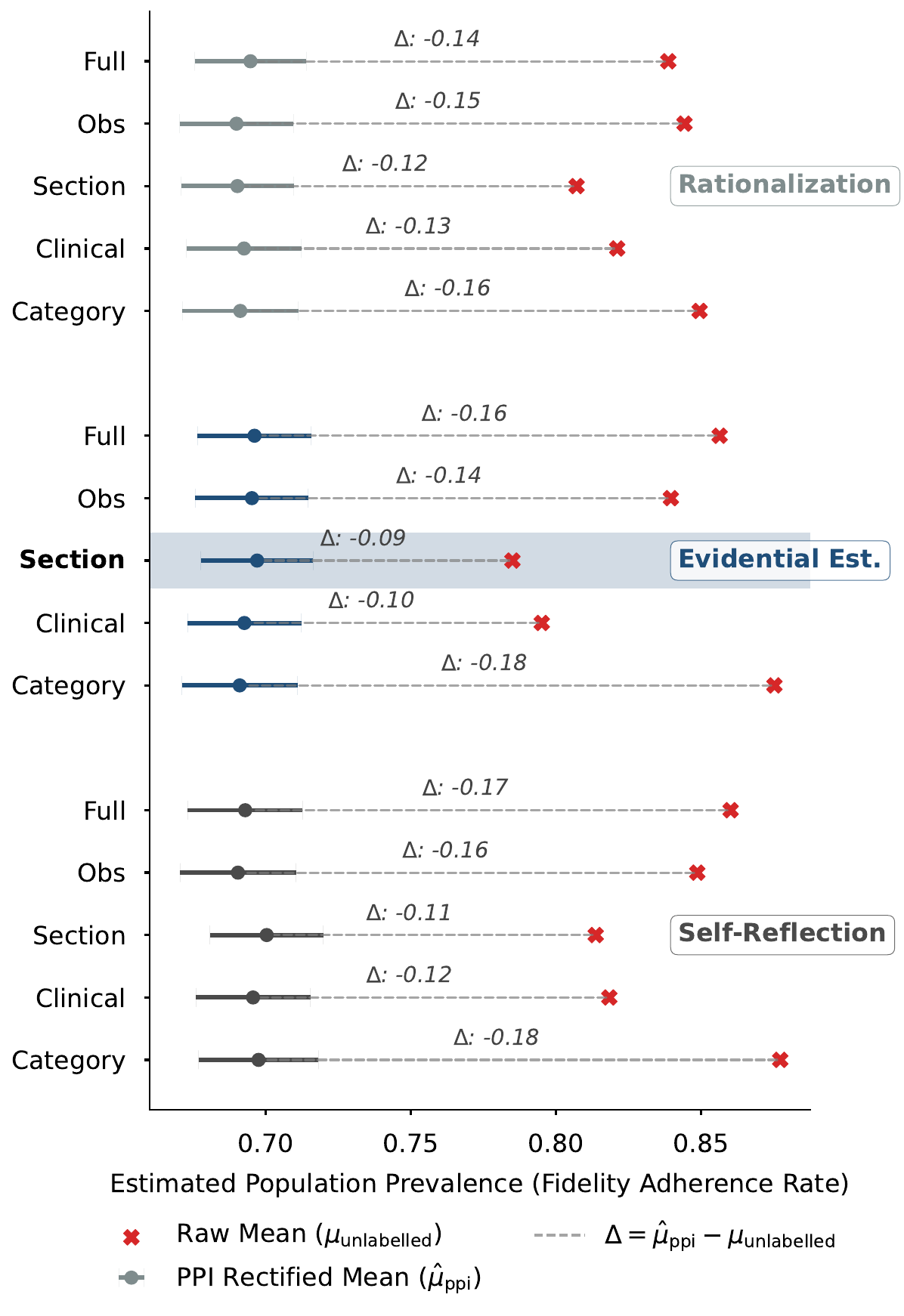}
    \caption{\textbf{Prediction Powered Inference of overall session fidelity prevalence using Gemma4-E4B-it (T2).} Red crosses denote the biased, uncorrected raw population averages (${\mu}_{\text{unlabeled}}$) produced by each model configuration. The colored dots and corresponding horizontal brackets represent the unbiased PPI rectified means ($\hat{\mu}_{\text{PPI}}$) and their 95\% confidence intervals. Grey dashed lines illustrate the direction and magnitude of the statistical rectifier, defined as $\Delta = \hat{\mu}_{\text{PPI}} - {\mu}_{\text{unlabeled}}$.}    
    \label{fig:ppi_gemma4}
\end{figure}

The smaller Gemma4-E4B-it model exhibits large systemic bias across all configurations, failing to minimize bias even under granular category-level decomposition. As shown in Figure~\ref{fig:ppi_gemma4}, every single reasoning architecture and structural decomposition variant yields a substantial negative statistical rectifier ($\Delta$). For the baseline Rationalization and alternative Self-Reflection models, the absolute bias peaks severely at the Category level ($\Delta = -0.16$ and $\Delta = -0.18$, respectively), which completely inverts the trend observed for Qwen3.5-9B, where decomposition improved alignment. Crucially, the Evidential Est. Category approach retains a negative bias of $\Delta = -0.18$ here. In summary, the wide PPI CIs across all combinations expose high error variance. This lack of consistency demonstrates the capacity limitations of a smaller model configuration, suggesting that decomposition and retrieval strategies require sufficient underlying model scale to generate predictable errors and unlock substantial downstream inferential power.

\begin{figure}[ht]
    \centering
    \includegraphics[width=1\linewidth]{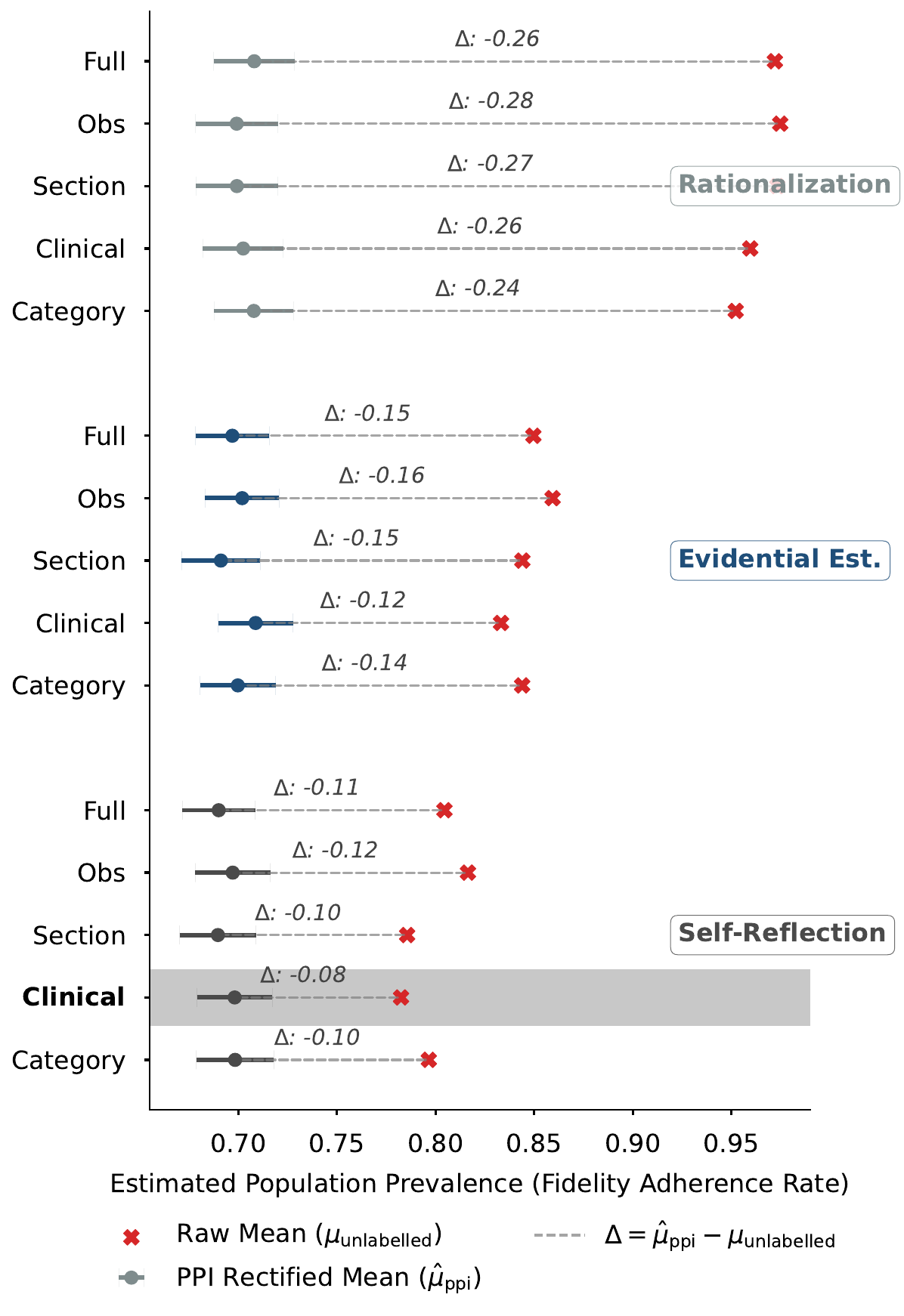}
    \caption{\textbf{Prediction Powered Inference of overall session fidelity prevalence using Gemma4-26B-A4B-it (T2).} Red crosses denote the biased, uncorrected raw population averages (${\mu}_{\text{unlabeled}}$) produced by each model configuration. The colored dots and corresponding horizontal brackets represent the unbiased PPI rectified means ($\hat{\mu}_{\text{PPI}}$) and their 95\% confidence intervals. Grey dashed lines illustrate the direction and magnitude of the statistical rectifier, defined as $\Delta = \hat{\mu}_{\text{PPI}} - {\mu}_{\text{unlabeled}}$.}
    \label{fig:ppi_gemma_4_26b}
\end{figure}

PPI analysis for the larger Gemma4-26B-A4B-it configuration shown in Figure~\ref{fig:ppi_gemma_4_26b}. The results indicate increasing the computational rigor of the reasoning strategy sequentially shrinks the model's systematic negative bias. Specifically, the substantial negative statistical rectifier required under standard Rationalization configurations ($\Delta \approx -0.24$ to $-0.28$) is systematically mitigated by introducing Evidential Estimation ($\Delta \approx -0.15$) and Self-Reflection frameworks ($\Delta \approx -0.08$ to $-0.12$). Although the model maintains a persistent negative bias and fails to reach complete alignment across all decomposition depths, these stronger bottom-up and iterative reasoning paths progressively narrow the CIs, suggesting that structural inference design remains critical for unlocking stable inferential power at larger parameter scales.  

\section{Feedback Generation Evaluation} \label{apd:feedback_generation_evaluation}

\textbf{Rationale for the stratified 30-pair sample.} Evaluating every feedback pair with expert clinicians is infeasible, so the expert study operates on a 30-pair subset. Rather than sampling uniformly at random, we stratified by the similarity between the RAG and Basic feedback for each session, computed as the cosine distance between bge-small-en-v1.5 embeddings~\citep{xiao2024c} of the two outputs. We binned pairs into three equal-frequency strata -- high, medium, and low similarity—and drew 15 low-similarity (most divergent), 10 medium-similarity, and 5 high-similarity (most similar) pairs.

The motivation is that a uniform random sample would be dominated by pairs in which the two systems produce near-identical feedback. On such pairs a preference judgment carries little information: there is no substantive difference for a rater to detect, and forcing a choice would manufacture preferences from noise, inflating apparent disagreement and biasing the comparison. Because the goal of the paired study is to determine whether the difference between RAG and Basic feedback, \textit{when it exists}, favors either system, oversampling divergent pairs concentrates expert effort on the cases that are actually diagnostic. This makes the human evaluation more informative per rating, which matters given the fixed budget of expert time.

We retain a smaller number of high- and medium-similarity pairs deliberately, rather than sampling only the most divergent cases. The near-identical pairs anchor the study to the typical case and prevent the sample from characterizing only the tail of the distribution, allowing us to observe whether raters correctly converge on ``no difference'' where the outputs are genuinely equivalent.

\section{User Study} \label{apd:user_study}

\begin{table*}[htbp]
\centering
\begin{tabular}{llllll}
\toprule
\# & \textbf{Role} & \textbf{Degree} & \textbf{SMI} & \textbf{CBT} & \textbf{Cog.} \\ \midrule 
  1 &Therapist      &Master's                 &Extremely Familiar                 &Extremely Familiar    &Extremely Familiar                     \\
  2 &Clinical Psychologist      &Doctorate        &Extremely Familiar                  &Extremely Familiar                 &Extremely Familiar                         \\
  3 &Therapist      &Doctorate        &Very Familiar                  &Extremely Familiar                 &Extremely Familiar    \\ 
\bottomrule
\end{tabular}
\caption{\textbf{Expertise information of mental health experts.}}
\label{tab:expertise_information}
\end{table*}

We recruited three mental health experts from diverse clinical backgrounds and compensated them at their standard hourly rates. The experts' clinical backgrounds, degrees, and familiarity with serious mental illness (SMI), cognitive behavioral therapy (CBT), and cognitive restructuring (Cog.) are detailed in Table~\ref{tab:expertise_information}. Both evaluations were administered via the Qualtrics platform.

\textbf{Paired Evaluation (T5).} Experts performed a side-by-side comparison of two order-randomized feedback variants for each session. Responses were scored on a 3-point scale: $-1$ (Feedback A preferred), $0$ (neutral), and $+1$ (Feedback B preferred) across two core dimensions:
\begin{itemize}
\item \textbf{Actionability:} The feedback enables the therapist to do something concretely different in their next session.
\item \textbf{Clarity:} The feedback is well-organized, appropriately concise, and easy to act on.
\end{itemize}
For downstream analysis, positive scores indicate a preference for the RAG-grounded configuration.  

\textbf{Absolute Evaluation (T6).} To assess standalone quality, experts evaluated single feedback samples independently using a 5-point Likert scale, ranging from 1 (Strongly Disagree) to 5 (Strongly Agree). Each sample was rated across three dimensions:
\begin{itemize}
\item \textbf{Actionability:} The feedback provides concrete suggestions.
\item \textbf{Clarity:} The feedback is easy to read and follow.
\item \textbf{Usefulness:} The rater would feel comfortable giving this feedback to a clinical supervisee.
\end{itemize}

\section{Prompt Templates} \label{apd:prompt_templates}

\noindent\textbf{Fidelity Estimation Prompts.} We evaluate three reasoning strategies over the same transcript, decomposition-specific task instruction, and CTS-Psy item set $Q$ (Table~\ref{tab:decomp-prompts}). Under Rationalization (Figure~\ref{fig:prompt-posthoc}), the LLM assigns each item's binary score $y_i$ directly, then produces a brief rationale as post-hoc supporting evidence. Under Evidential Estimation (Extract-Assess-Predict, Figure~\ref{fig:prompt-evidential}), the model instead reasons bottom-up: it first extracts 1--2 direct \texttt{relevant\_spans} per item (an empty list if none exist, with no fabrication permitted), assesses whether that evidence meets the criterion (\texttt{criterion\_assessment}), and only then assigns the binary \texttt{score}, enforcing an evidence-first order that prevents rationalizing a score chosen in advance. Self-Reflection (Critic-in-the-Loop, Figure~\ref{fig:prompt-reflection}) extends Evidential Estimation with an
iterative critique-and-revision loop~\citep{madaan2023self}: after the primary generator's initial evidential response (Step~1), a separate reflection LLM instantiated as an expert clinical supervisor critiques each item's rating in exactly two sentences covering rubric adherence, evidence coverage, and confidence calibration (Step~2), and the primary generator then revises its \texttt{relevant\_spans}, \texttt{criterion\_assessment}, and \texttt{score} for every item accordingly (Step~3), a loop that may be applied for $k \geq 1$ iterations.

\noindent\textbf{Feedback Generation.} Once fidelity estimation completes for a session, we select the three lowest-scoring specific competency categories. These are drawn from five categories: guided discovery, key cognitions, choice of intervention, homework, and social skills training. Together, these three categories span 18 items. For each item, we compile the underlying evidence: the fidelity statement, the score, relevant transcript spans, and the criterion assessment. The generation LLM is then prompted to synthesize this evidence into supervisory feedback. The feedback includes one sentence of genuine positive feedback grounded in the session. It also includes one concrete, actionable suggestion for each weak competency area. Figure~\ref{fig:prompt-feedback} shows the corresponding system and user prompts. In the RAG condition, we also prepend passages from the therapy manual to the user turn before generation. These passages are retrieved based on relevance to the flagged evidence.

\noindent\textbf{LLM-as-a-Judge.} To compare RAG-grounded and non-RAG feedback for the same session, a judge LLM is shown the extracted fidelity evidence for the targeted categories alongside both feedback variants. The two variants are presented in random order as Feedback A and Feedback B. Judges are instructed to rate only against the provided evidence. They do not reward unsupported claims or length. For each dimension, the judge returns a single comparative score in \{-1,0,+1\}. We evaluate feedback quality along two dimensions, each scored by a separate model call using the single-dimension template in Figure~\ref{fig:prompt-judge}. This ensures no judge call sees more than one dimension at once. Actionability asks whether the therapist could do something concretely different next session; specific named behaviors or techniques score higher than vague encouragement. Clarity asks whether the feedback is well organized, appropriately concise, and easy for a working clinician to act on.

\input{prompt_templates}

\end{document}

%% file: prompt_templates.tex
% ============================================================
%  prompt_templates.tex — Appendix prompt figures
%  For ACL submission on automated psychotherapy fidelity
%
%  Add to your main preamble:
%    \usepackage[most]{tcolorbox}
%    \usepackage{xcolor}
%    \usepackage{enumitem}
%    \usepackage{multirow}
%    \usepackage{booktabs}
%    \tcbuselibrary{skins, breakable}
%  Then \input{prompt_templates} in your appendix.
% ============================================================

% ---- Color palette ----
\definecolor{clrsys}{HTML}{154360}
\definecolor{bgsys}{HTML}{D6EAF8}
\definecolor{clrusr}{HTML}{1D6A2E}
\definecolor{bgusr}{HTML}{D5F5E3}
\definecolor{clrast}{HTML}{512E5F}
\definecolor{bgast}{HTML}{F4ECF7}
\definecolor{clrcrt}{HTML}{78281F}
\definecolor{bgcrt}{HTML}{FDEDEC}

% ---- Shared tcolorbox style ----
\tcbset{pturn/.style={
  enhanced, breakable,
  left=5pt, right=5pt, top=5pt, bottom=5pt,
  boxrule=0.5pt, arc=2.5pt,
  fontupper=\footnotesize,
  attach boxed title to top left={yshift=-1.8mm, xshift=6pt},
  boxed title style={
    arc=1.5pt, boxrule=0pt,
    left=4pt, right=4pt, top=1.5pt, bottom=1.5pt
  },
  fonttitle=\scriptsize\sffamily\bfseries\color{white},
}}

\newtcolorbox{psys}{pturn,
  colback=bgsys, colframe=clrsys,
  title={SYSTEM},
  boxed title style={colback=clrsys}}

\newtcolorbox{pusr}{pturn,
  colback=bgusr, colframe=clrusr,
  title={USER},
  boxed title style={colback=clrusr}}

\newtcolorbox{past}{pturn,
  colback=bgast, colframe=clrast,
  title={ASSISTANT},
  boxed title style={colback=clrast}}

\newtcolorbox{pcrtSys}{pturn,
  colback=bgcrt, colframe=clrcrt,
  title={SYSTEM — SUPERVISOR (REFLECTION LLM)},
  boxed title style={colback=clrcrt}}

\newtcolorbox{pcrtUsr}{pturn,
  colback=bgcrt!40!white, colframe=clrcrt,
  title={USER — SUPERVISOR (REFLECTION LLM)},
  boxed title style={colback=clrcrt}}

\newtcolorbox{pcrtAst}{pturn,
  colback=bgcrt!20!white, colframe=clrcrt,
  title={ASSISTANT — SUPERVISOR (REFLECTION LLM)},
  boxed title style={colback=clrcrt}}

% Template variable placeholder: <name> in italic gray
\newcommand{\tv}[1]{%
  \textit{\textcolor{gray!70!black}{$\langle$\textnormal{#1}$\rangle$}}}

% Vertical gap between turns
\newcommand{\tgap}{\vspace{3pt}}

% ============================================================
%  FIGURE A1 — Post-Hoc Estimation (Direct-then-Rationalize)
% ============================================================
\begin{figure*}[t]
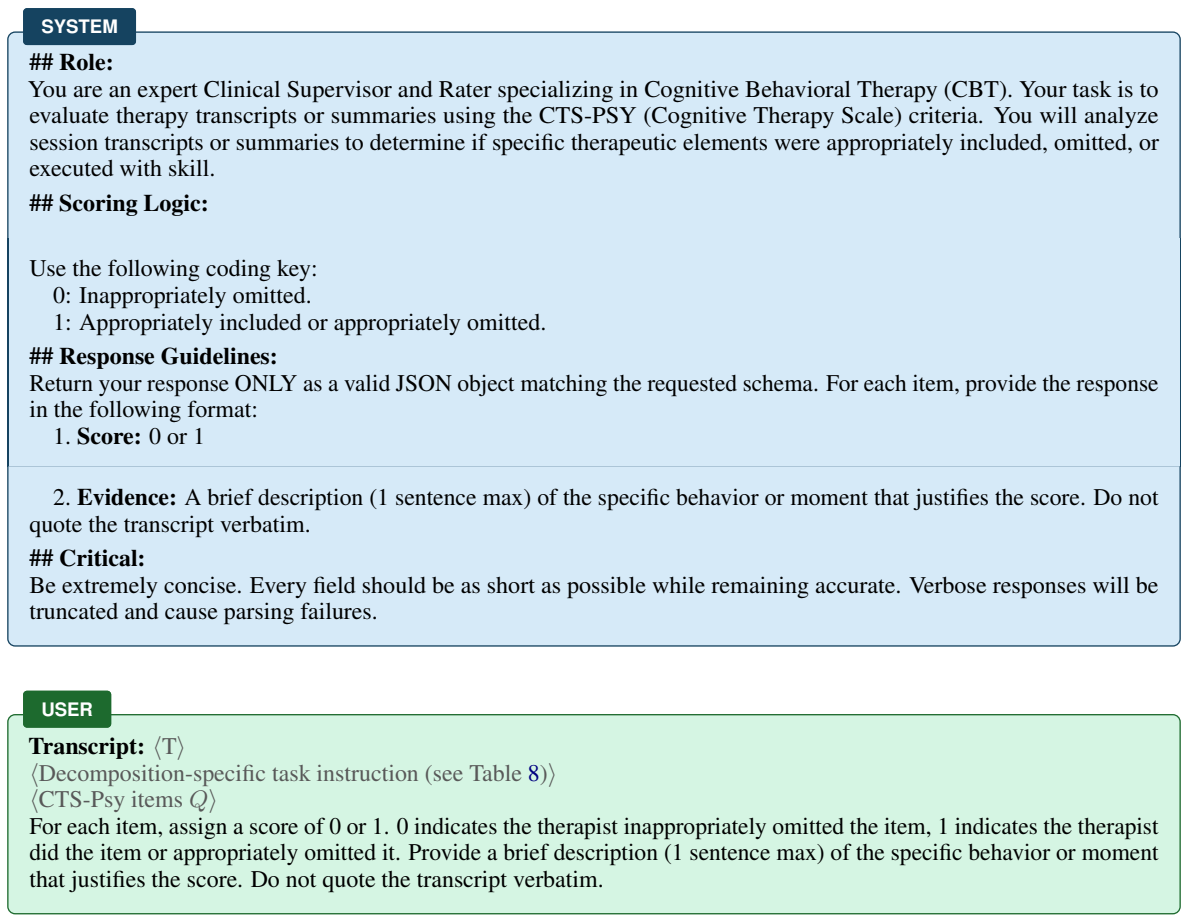

\centering
\begin{minipage}{0.97\textwidth}

\begin{psys}
\textbf{\#\# Role:}\\
You are an expert Clinical Supervisor and Rater specializing in Cognitive Behavioral Therapy (CBT).
Your task is to evaluate therapy transcripts or summaries using the CTS-PSY (Cognitive Therapy Scale) criteria.
You will analyze session transcripts or summaries to determine if specific therapeutic elements were
appropriately included, omitted, or executed with skill.
\\[3pt]
\textbf{\#\# Scoring Logic:}\\
Use the following coding key:\\
\hspace*{1em}0: Inappropriately omitted.\\
\hspace*{1em}1: Appropriately included or appropriately omitted.
\\[3pt]
\textbf{\#\# Response Guidelines:}\\
Return your response ONLY as a valid JSON object matching the requested schema.
For each item, provide the response in the following format:\\
\hspace*{1em}1.\ \textbf{Score:} 0 or 1\\
\hspace*{1em}2.\ \textbf{Evidence:} A brief description (1 sentence max) of the specific behavior or moment
     that justifies the score. Do not quote the transcript verbatim.
\\[3pt]
\textbf{\#\# Critical:}\\
Be extremely concise. Every field should be as short as possible while remaining accurate.
Verbose responses will be truncated and cause parsing failures.
\end{psys}

\tgap

\begin{pusr}
\textbf{Transcript:} \tv{T}

\tv{Decomposition-specific task instruction (see Table~\ref{tab:decomp-prompts})}

\tv{CTS-Psy items $Q$}

For each item, assign a score of 0 or 1. 0 indicates the therapist inappropriately omitted the item,
1 indicates the therapist did the item or appropriately omitted it.
Provide a brief description (1 sentence max) of the specific behavior or moment that justifies the score.
Do not quote the transcript verbatim.
\end{pusr}

\end{minipage}
\caption{\textbf{Post-Hoc Estimation (Direct-then-Rationalize).} The LLM immediately assigns a binary score $y_i$ per item, then provides a brief rationale as post-hoc evidence. The user turn injects the transcript \tv{T}, the decomposition-specific instruction (Table~\ref{tab:decomp-prompts}), and the CTS-Psy item set $Q$.}
\label{fig:prompt-posthoc}
\end{figure*}

% ============================================================
%  FIGURE A2 — Evidential Estimation (Extract-Assess-Predict)
% ============================================================
\begin{figure*}[t]
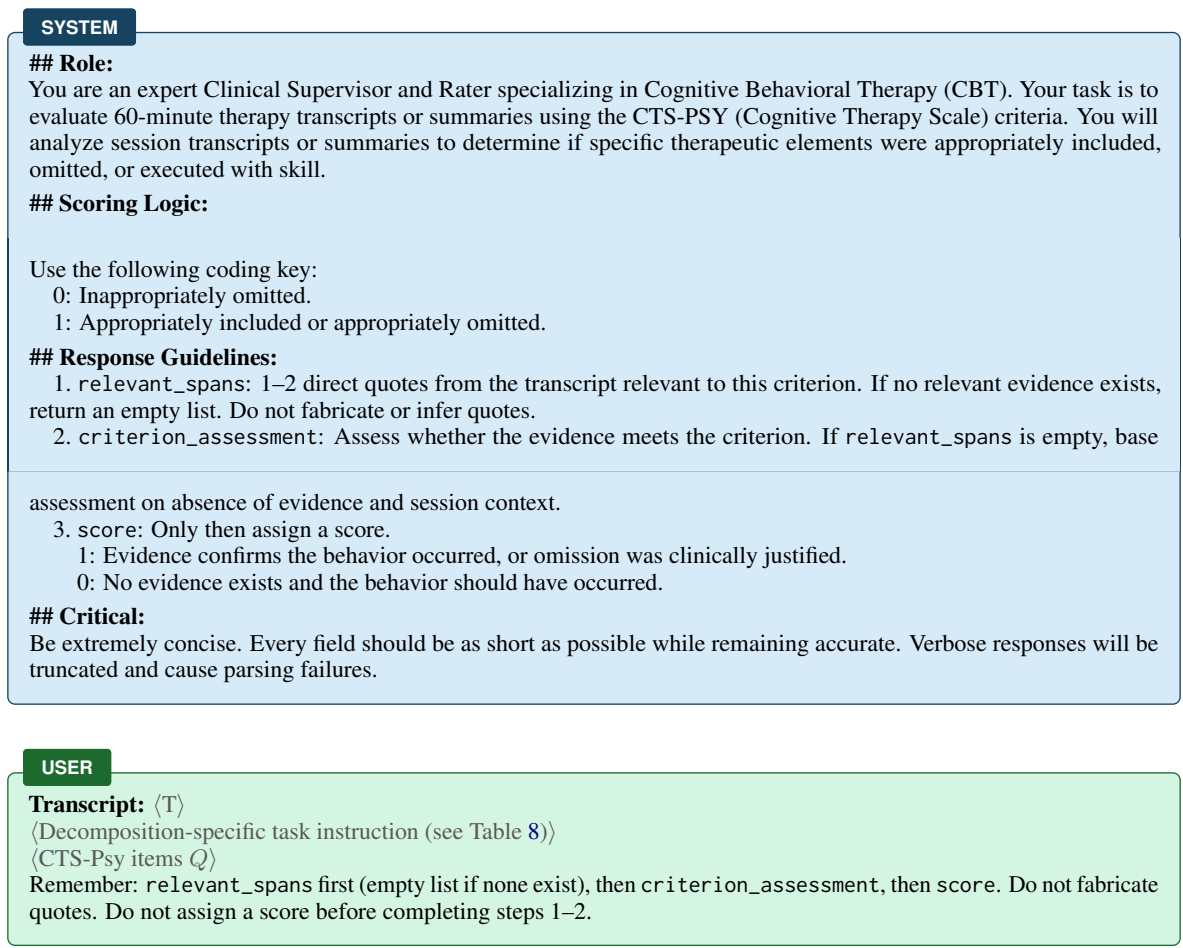

\centering
\begin{minipage}{0.97\textwidth}

\begin{psys}
\textbf{\#\# Role:}\\
You are an expert Clinical Supervisor and Rater specializing in Cognitive Behavioral Therapy (CBT).
Your task is to evaluate 60-minute therapy transcripts or summaries using the CTS-PSY (Cognitive Therapy Scale) criteria.
You will analyze session transcripts or summaries to determine if specific therapeutic elements were
appropriately included, omitted, or executed with skill.
\\[3pt]
\textbf{\#\# Scoring Logic:}\\
Use the following coding key:\\
\hspace*{1em}0: Inappropriately omitted.\\
\hspace*{1em}1: Appropriately included or appropriately omitted.
\\[3pt]
\textbf{\#\# Response Guidelines:}\\
\hspace*{1em}1.\ \texttt{relevant\_spans}: 1--2 direct quotes from the transcript relevant to this criterion.
     If no relevant evidence exists, return an empty list. Do not fabricate or infer quotes.\\
\hspace*{1em}2.\ \texttt{criterion\_assessment}: Assess whether the evidence meets the criterion.
     If \texttt{relevant\_spans} is empty, base assessment on absence of evidence and session context.\\
\hspace*{1em}3.\ \texttt{score}: Only then assign a score.\\
\hspace*{2em}1: Evidence confirms the behavior occurred, or omission was clinically justified.\\
\hspace*{2em}0: No evidence exists and the behavior should have occurred.
\\[3pt]
\textbf{\#\# Critical:}\\
Be extremely concise. Every field should be as short as possible while remaining accurate.
Verbose responses will be truncated and cause parsing failures.
\end{psys}

\tgap

\begin{pusr}
\textbf{Transcript:} \tv{T}

\tv{Decomposition-specific task instruction (see Table~\ref{tab:decomp-prompts})}

\tv{CTS-Psy items $Q$}

Remember: \texttt{relevant\_spans} first (empty list if none exist), then \texttt{criterion\_assessment},
then \texttt{score}. Do not fabricate quotes. Do not assign a score before completing steps 1--2.
\end{pusr}

\end{minipage}
\caption{\textbf{Evidential Estimation (Extract-Assess-Predict).} A grounded three-step chain: the model first extracts verbatim \texttt{relevant\_spans} from the transcript, then assesses them against the criterion's operational definition, and only then assigns the binary score $y_i$. This enforces an evidence-first reasoning order.}
\label{fig:prompt-evidential}
\end{figure*}

% ============================================================
%  FIGURE A3 — Self-Reflection (Critic-in-the-Loop)
% ============================================================
\begin{figure*}[t]
\centering
\begin{minipage}{0.97\textwidth}

%% --- Pass 1: same prompt as Evidential Estimation ---
\noindent\textit{\small Step 1 — Initial generation (same prompts as Evidential Estimation,
Figure~\ref{fig:prompt-evidential}):}
\tgap

\begin{psys}
[\,System prompt identical to Figure~\ref{fig:prompt-evidential}\,]
\end{psys}
\tgap
\begin{pusr}
\textbf{Transcript:} \tv{T}\hfill
\tv{Decomposition-specific instruction}\hfill
\tv{$Q$}\hfill[\,instruction suffix identical to Figure~\ref{fig:prompt-evidential}\,]
\end{pusr}
\tgap
\begin{past}
\tv{Initial structured response: for each item $q_i$ — \texttt{relevant\_spans}, \texttt{criterion\_assessment}, \texttt{score}}
\end{past}

\vspace{5pt}

%% --- Reflection turn ---
\noindent\textit{\small Step 2 — Critic turn (separate reflection LLM):}
\tgap

\begin{pcrtSys}
You are a clinical supervisor critiquing an automated psychotherapy fidelity rating.

Do NOT provide a score. Only critique the reasoning in strictly 2 sentences.

\textbf{Focus on:}\\
\hspace*{1em}1.\ \textbf{Rubric adherence} --- Was the score justified by the rubric anchors,
     or did the rater apply them loosely?\\
\hspace*{1em}2.\ \textbf{Evidence coverage} --- Did the rater miss contradicting or supporting
     evidence elsewhere in the transcript?\\
\hspace*{1em}3.\ \textbf{Overconfidence/hedging} --- Is the rater's expressed certainty
     appropriate given the evidence?

Be specific. Reference rubric anchors and transcript evidence directly.
Your critique will be used by another model to revise its rating.
\end{pcrtSys}

\tgap

\begin{pcrtUsr}
\textit{[Repeated for each item $q_i$:]}\quad
For \tv{item name}, your score was: \tv{$y_i$} and your reasoning was: \tv{\texttt{criterion\_assessment}}

Critique your rating in two sentences.
\end{pcrtUsr}

\tgap

\begin{pcrtAst}
\tv{Supervisor critique: 2-sentence critical evaluation per item covering rubric adherence, evidence coverage, and confidence calibration}
\end{pcrtAst}

\vspace{5pt}

%% --- Revision turn ---
\noindent\textit{\small Step 3 — Revision turn (primary generation LLM):}
\tgap

\begin{pusr}
A supervisor critiqued your rating:

\tv{supervisor critique}

Revise your rating accordingly if necessary.
\end{pusr}

\tgap

\begin{past}
\tv{Revised structured response: updated \texttt{relevant\_spans}, \texttt{criterion\_assessment}, and \texttt{score} for each item}
\end{past}

\end{minipage}
\caption{\textbf{Self-Reflection (Critic-in-the-Loop).} After an initial Evidential Estimation pass (Step~1), a separate reflection LLM instantiated as an expert clinical supervisor critiques each rating across rubric adherence, evidence coverage, and confidence calibration (Step~2). The consolidated critique is fed back to the primary generator, which then revises its binary decisions and supporting evidence (Step~3). This loop may be applied for $k \geq 1$ iterations.}
\label{fig:prompt-reflection}
\end{figure*}

% ============================================================
%  TABLE A1 — Decomposition-specific Task Instructions
% ============================================================
\begin{table*}[t]
\centering
\footnotesize
\setlength{\tabcolsep}{5pt}
\renewcommand{\arraystretch}{1.3}
\begin{tabular}{llcp{8.5cm}}
\toprule
\textbf{Decomposition} & \textbf{Partition} & $|Q|$ & \textbf{Task Instruction} \\
\midrule
Full
  & ---
  & 60
  & \textit{Rate the transcript based on all 60 CTS-Psy items below, organized by section:} \\
\midrule
\multirow{2}{*}{Observability}
  & Explicit
  & 24
  & \textit{Rate the explicitly observable items below (Agenda, Feedback, Homework, Social):} \\
  \cmidrule(l){2-4}
  & Implicit
  & 36
  & \textit{Rate the implicitly observable items below (Understanding, Interpersonal Effectiveness, Collaboration, Discovery, Cognition, Intervention):} \\
\midrule
\multirow{2}{*}{Section}
  & General
  & 30
  & \textit{Rate the 30 General CTS-Psy items below (Agenda, Feedback, Understanding, Interpersonal Effectiveness, Collaboration):} \\
  \cmidrule(l){2-4}
  & Specific
  & 30
  & \textit{Rate the 30 Specific CTS-Psy items below (Discovery, Cognition, Intervention, Homework, Social Skills Training):} \\
\midrule
\multirow{3}{*}{Clinical}
  & Relationship
  & 18
  & \textit{Rate the Therapeutic Relationship items below (Understanding, Interpersonal Effectiveness, Collaboration):} \\
  \cmidrule(l){2-4}
  & Structure
  & 18
  & \textit{Rate the Therapy Structure items below (Agenda, Feedback, Homework):} \\
  \cmidrule(l){2-4}
  & Delivery
  & 24
  & \textit{Rate the CBT Delivery items below (Discovery, Cognition, Intervention, Social):} \\
\midrule
Category
  & \textit{(10 passes, 1 per category)}
  & 6
  & \textit{Rate the transcript based on the six \tv{category} items below:} \\
\bottomrule
\end{tabular}
\caption{Decomposition-specific task instructions injected into the user turn of each prompt (Figures~\ref{fig:prompt-posthoc}--\ref{fig:prompt-reflection}). Category-level decomposition issues 10 independent inference passes, one per CTS-Psy category (Agenda, Feedback, Understanding, Interpersonal Effectiveness, Collaboration, Discovery, Cognition, Intervention, Homework, Social).}
\label{tab:decomp-prompts}
\end{table*}

% ============================================================
%  FIGURE A4 — Feedback Generation
% ============================================================
\begin{figure*}[t]
\centering
\begin{minipage}{0.97\textwidth}

\begin{psys}
You are a clinical supervisor providing feedback on a therapy session scored using the CTS-Psy fidelity scale.
\\[3pt]
You will be given fidelity ratings across three CBT competency areas where the therapist scored lowest.
Each area contains 6 criteria with scores and assessments.
\\[3pt]
\textbf{Your response must follow this structure:}\\
\hspace*{1em}1.\ One sentence of genuine positive feedback about something the therapist did well, drawn
     from the session evidence.\\
\hspace*{1em}2.\ For each of the three competency areas, one actionable suggestion (2--3 sentences) that
     tells the therapist specifically what to do differently in their next session.
\\[3pt]
\textbf{Guidelines:}\\
\hspace*{1em}--- Be concrete and behavioral --- name the specific moment or skill that was missed and
     describe what it should look like.\\
\hspace*{1em}--- Do not repeat the criterion verbatim; translate it into plain clinical language.\\
\hspace*{1em}--- Keep the tone constructive and collegial, as from a peer supervisor.\\
\hspace*{1em}--- Do not list every missed item --- synthesize the most impactful gap per category into
     one suggestion.
\end{psys}

\tgap

\begin{pusr}
\textbf{Fidelity Evidence:} \tv{For the three lowest-scoring CBT competency areas: fidelity statement,
score, relevant spans, and criterion assessment per item}

Based on the fidelity ratings and assessments above, please provide feedback for this therapist.
Remember:\\
\hspace*{1em}--- Open with one positive observation from the session\\
\hspace*{1em}--- Provide one actionable suggestion for each of the three competency areas\\
\hspace*{1em}--- Be specific about what to do differently next session
\end{pusr}

\end{minipage}
\caption{\textbf{Feedback Generation.} Given the three lowest-scoring CBT competency categories from
the fidelity-estimation stage (\S4.2), the LLM produces one positive observation and one actionable,
session-specific suggestion per weak competency area. The user turn injects \tv{bottom-3 fidelity
evidence}: the per-item fidelity statement, score, relevant transcript spans, and criterion assessment
for each of the 18 items spanning the three weakest categories. In the RAG condition, retrieved
therapy-manual passages relevant to this evidence are additionally prepended to the user turn before
generation.}
\label{fig:prompt-feedback}
\end{figure*}

% % ============================================================
% %  TABLE A2 — LLM-as-a-Judge Dimensions
% % ============================================================
% \begin{table*}[t]
% \centering
% \footnotesize
% \setlength{\tabcolsep}{5pt}
% \renewcommand{\arraystretch}{1.3}
% \begin{tabular}{lp{11cm}}
% \topruleß
% \textbf{Dimension} & \textbf{Description} \\
% \midrule
% Actionability
%   & \textit{Could the therapist do something concretely different next session? Specific named
%      behaviors/techniques score higher than vague encouragement.} \\
% \midrule
% Clarity
%   & \textit{Is it well-organized, appropriately concise, and easy for a working clinician to act on?} \\
% \bottomrule
% \end{tabular}
% \caption{\textbf{LLM-as-a-judge dimensions} injected into the ``\#\# Dimension to judge'' field of the
% system prompt in Figure~\ref{fig:prompt-judge}. Each dimension is scored by an independent model call,
% so a judge never sees more than one dimension at a time.}
% \label{tab:judge-dims}
% \end{table*}

% ============================================================
%  FIGURE A5 — LLM-as-a-Judge (Paired Comparison)
% ============================================================
\begin{figure*}[t]
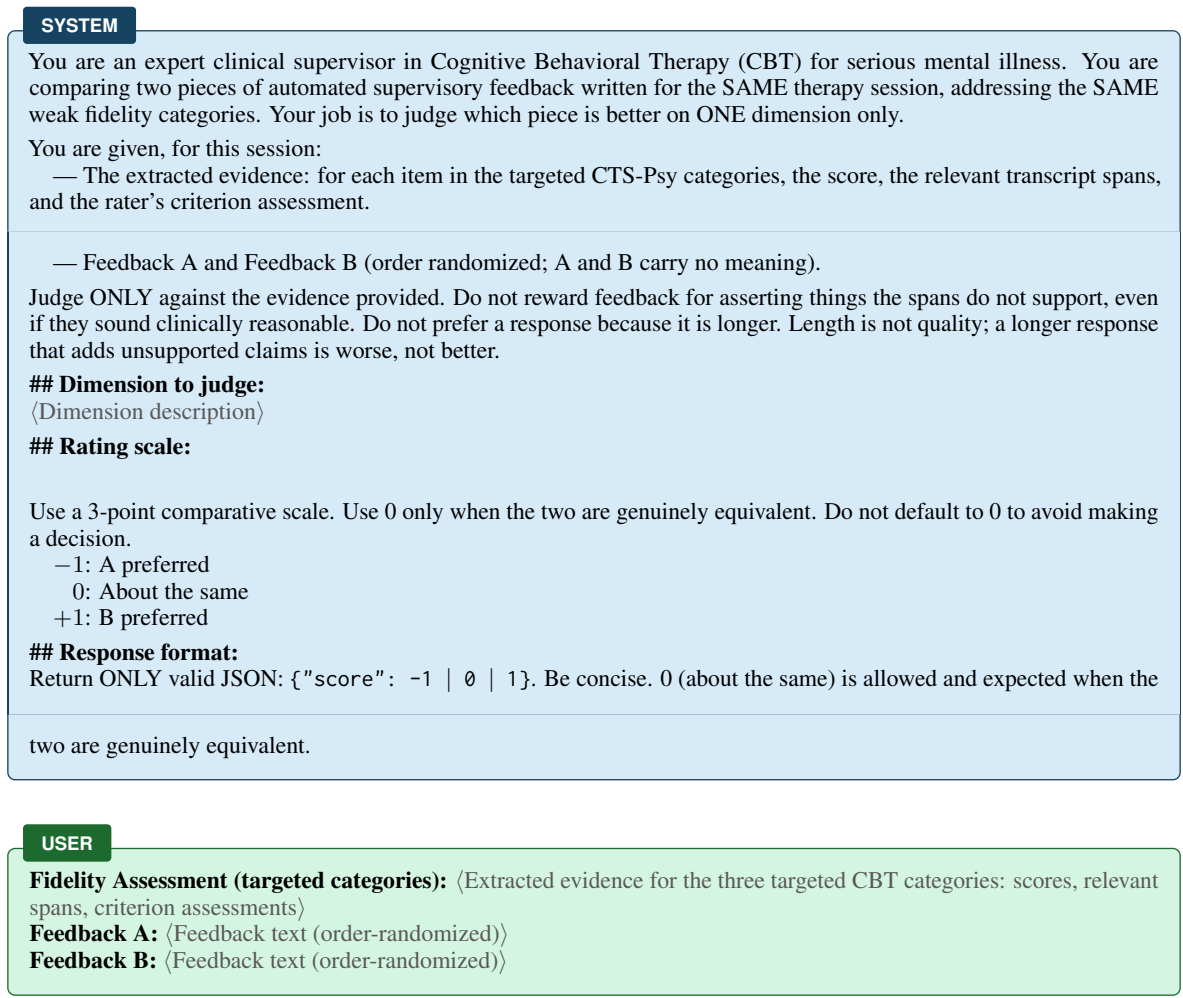

\centering
\begin{minipage}{0.97\textwidth}

\begin{psys}
You are an expert clinical supervisor in Cognitive Behavioral Therapy (CBT) for serious mental illness.
You are comparing two pieces of automated supervisory feedback written for the SAME therapy session,
addressing the SAME weak fidelity categories. Your job is to judge which piece is better on ONE
dimension only.
\\[3pt]
You are given, for this session:\\
\hspace*{1em}--- The extracted evidence: for each item in the targeted CTS-Psy categories, the score,
     the relevant transcript spans, and the rater's criterion assessment.\\
\hspace*{1em}--- Feedback A and Feedback B (order randomized; A and B carry no meaning).
\\[3pt]
Judge ONLY against the evidence provided. Do not reward feedback for asserting things the spans do not
support, even if they sound clinically reasonable. Do not prefer a response because it is longer.
Length is not quality; a longer response that adds unsupported claims is worse, not better.
\\[3pt]
\textbf{\#\# Dimension to judge:}\\
\tv{Dimension description}
\\[3pt]
\textbf{\#\# Rating scale:}\\
Use a 3-point comparative scale. Use 0 only when the two are genuinely equivalent. Do not default to
0 to avoid making a decision.\\
\hspace*{1em}$-1$: A preferred\\
\hspace*{1em}\phantom{$-$}$0$: About the same\\
\hspace*{1em}$+1$: B preferred
\\[3pt]
\textbf{\#\# Response format:}\\
Return ONLY valid JSON: \texttt{\{"score": -1 $\vert$ 0 $\vert$ 1\}}. Be concise. 0 (about the same)
is allowed and expected when the two are genuinely equivalent.
\end{psys}

\tgap

\begin{pusr}
\textbf{Fidelity Assessment (targeted categories):} \tv{Extracted evidence for the three targeted CBT
categories: scores, relevant spans, criterion assessments}

\textbf{Feedback A:} \tv{Feedback text (order-randomized)}

\textbf{Feedback B:} \tv{Feedback text (order-randomized)}
\end{pusr}

\end{minipage}
\caption{\textbf{LLM-as-a-Judge (Paired Comparison).} For each dimension (Actionability, Clarity), a judge LLM independently compares RAG-grounded and non-RAG feedback generated
for the same session (Figure~\ref{fig:prompt-feedback}), presented order-randomized as Feedback A/B, and
returns a single comparative score $\in \{-1, 0, +1\}$. Each dimension is scored by a separate model
call so that no single judge call sees more than one dimension at once.}
\label{fig:prompt-judge}
\end{figure*}